\documentclass[10pt,journal,compsoc]{IEEEtran}

\usepackage{mathtools}
\usepackage{adjustbox}
\usepackage{amsmath}
\usepackage{amssymb}
\usepackage{amsthm}
\usepackage{amsfonts}
\usepackage{algorithm2e}
\usepackage{graphicx}
\usepackage{xcolor}
\usepackage{float}
\usepackage{hyperref}

\usepackage{url}

\usepackage{multirow}
\usepackage{tikz}
\usepackage{pgfplots}

\usepackage{CJKutf8}
\usepackage{adjustbox}

\definecolor{bblue}{HTML}{4F81BD}
\definecolor{rred}{HTML}{C0504D}
\definecolor{ggreen}{HTML}{9BBB59}
\definecolor{auburn}{HTML}{6D351A}
\definecolor{atangerine}{HTML}{FF9966}
\definecolor{ppurple}{HTML}{9F4C7C}

\usepackage{framed,multirow}
\usepackage{pifont}
\usepackage{diagbox}
\newcommand{\cmark}{\ding{51}}%
\newcommand{\xmark}{\ding{55}}%

\usepackage{lipsum}
\usepackage{graphicx}
\ifCLASSOPTIONcompsoc
    \usepackage[caption=false, font=normalsize, labelfont=sf, textfont=sf]{subfig}
\else
\usepackage[caption=false, font=footnotesize]{subfig}
\fi

\newcommand\copyrighttext{%
  \footnotesize \textcopyright 2012 IEEE. Personal use of this material is permitted.
  Permission from IEEE must be obtained for all other uses, in any current or future 
  media, including reprinting/republishing this material for advertising or promotional 
  purposes, creating new collective works, for resale or redistribution to servers or 
  lists, or reuse of any copyrighted component of this work in other works. 
  DOI: \href{<http://tex.stackexchange.com>}{<DOI No.>}}
\newcommand\copyrightnotice{%
\begin{tikzpicture}[remember picture,overlay]
\node[anchor=south,yshift=10pt] at (current page.south) {\fbox{\parbox{\dimexpr\textwidth-\fboxsep-\fboxrule\relax}{\copyrighttext}}};
\end{tikzpicture}%
}

\newcommand{\rvect}{\mbox{\bf r}}

\newcommand{\tvect}{\mbox{\bf t}}

\newcommand{\xvect}{\mbox{\bf x}}

\newcommand{\zvect}{\mbox{\bf z}}

\ifCLASSOPTIONcompsoc
  \usepackage[nocompress]{cite}
\else
  \usepackage{cite}
\fi

\ifCLASSINFOpdf
\else
\fi

\begin{document}

\title{Extremely Fine-Grained Visual Classification over Resembling Glyphs in the Wild}
%
%
%
%

 
\author{Fares~Bougourzi, Fadi~Dornaika, and Chongsheng~Zhang\IEEEauthorrefmark{1}
\IEEEcompsocitemizethanks{\IEEEcompsocthanksitem Fares Bougourzi and  Chongsheng Zhang are with the School of Computer and Information Engineering, Henan University, China. Pr. Fadi Dornaika is with UPV/EHU and IKERBASQUE, Basque Foundation  for Science, Spain E-mail: faresbougourzi@gmail.com, fdornaika@gmail.com, cszhang@ieee.org. 

\IEEEcompsocthanksitem \IEEEauthorrefmark{1} denotes corresponding author.}
}


\IEEEtitleabstractindextext{%
\begin{abstract}
Text recognition in the wild is an important technique for digital maps and urban scene understanding, in which the natural resembling properties between glyphs is one of the major reasons that lead to wrong recognition results. To address this challenge, we introduce two extremely fine-grained visual recognition benchmark datasets that contain very challenging resembling glyphs (characters/letters) in the wild to be  distinguished. Moreover, we propose a simple yet effective two-stage contrastive learning approach to the extremely fine-grained recognition task of resembling glyphs discrimination. In the first stage, we utilize supervised contrastive learning to leverage label information to warm-up the backbone network. In the second stage, we introduce CCFG-Net, a network architecture that integrates classification and contrastive learning in both Euclidean and Angular spaces, in which contrastive learning is applied in both supervised learning and pairwise discrimination manners to enhance the model's feature representation capability. Overall, our proposed approach effectively exploits the complementary strengths of contrastive learning and classification, leading to improved recognition performance on the resembling glyphs. Comparative evaluations with state-of-the-art fine-grained classification approaches under both Convolutional Neural Network (CNN) and Transformer backbones demonstrate the superiority of our proposed method.

\end{abstract}

\begin{IEEEkeywords}
 Low-shot Object Recognition, Resembling Characters, Resembling Letters, Fine-grained Visual Classification.
\end{IEEEkeywords}}

\maketitle
\copyrightnotice

\IEEEdisplaynontitleabstractindextext

%
\IEEEpeerreviewmaketitle

\IEEEraisesectionheading{\section{Introduction}\label{sec:introduction}}

\IEEEPARstart{T}{he}
advent of writing thousands of years ago enabled the preservation of knowledge across generations. Early ancient Chinese characters, such as Oracle Bone Inscriptions, were hieroglyphs. After that, pictophonetic characters have been formed by combining semantic symbols which hint at the character's meaning and phonetic symbols which suggest its pronunciation. Noways, more than 90\% of the Chinese characters are pictophonetic \cite{boltz1986early, boltz1994origin}, making it  difficult to infer their meanings solely from  the glyphs or radicals.  

\begin{figure}
\centering
\includegraphics[width=9cm]{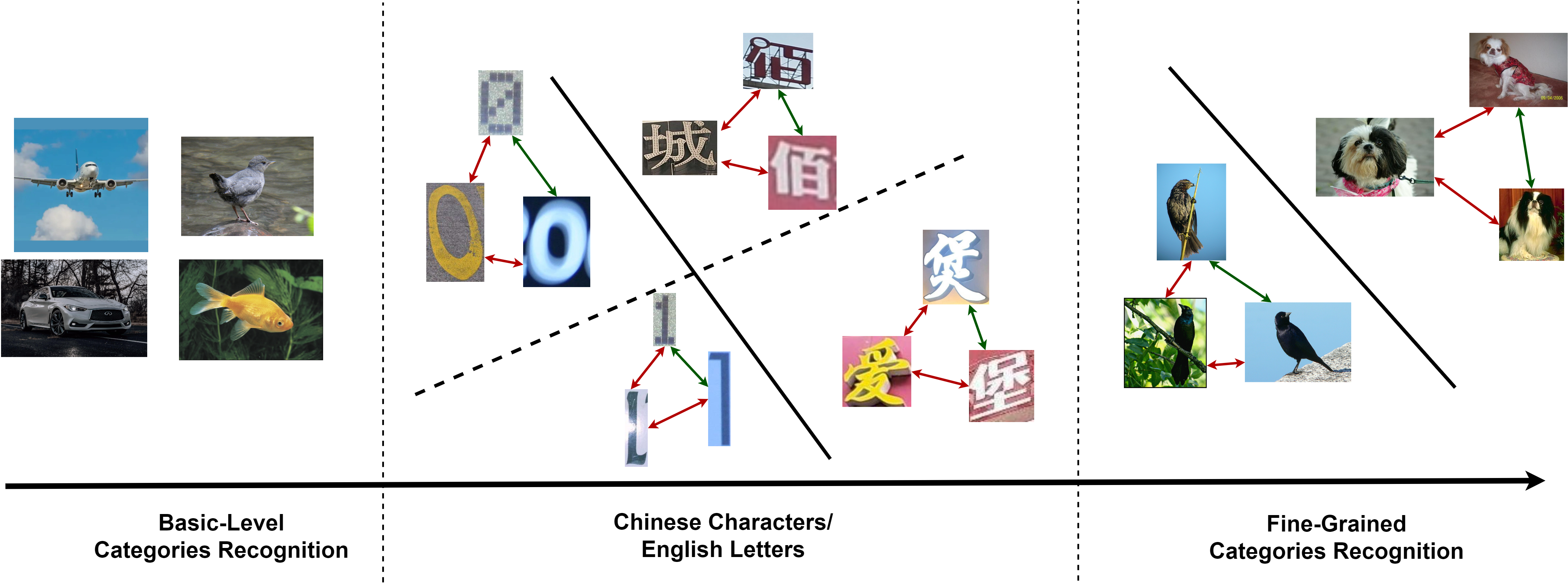}
\caption{An illustration showing the differences between conventional fine-grained object recognition  (e.g., bird species) and resembling glyphs discrimination. The former typically have common semantic parts, which are not available in the latter, making it a significantly challenging and extremely fine-grained visual recognition task.} 
\label{fig:CLGF}
\end{figure}

\begin{figure}[h!]
\centering
\resizebox{0.35\textwidth}{!}
{
\includegraphics[page=1,width=0.9\textwidth]{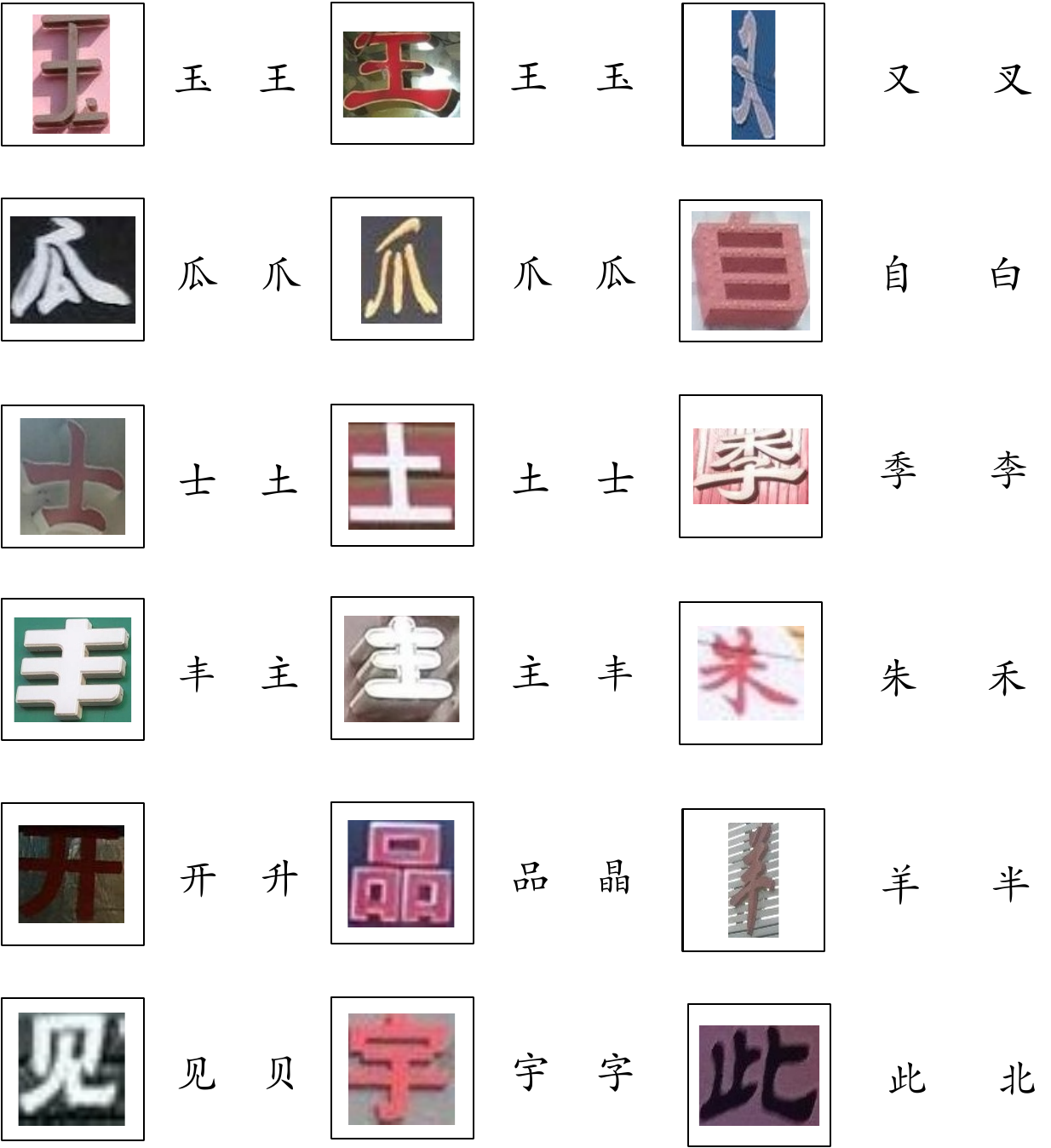}}
\caption{Examples of Chinese Resembling Characters. For each scene character image in the figure, we provide both the ground truth character/class and the predicted class, respectively.}
\label{rcc-example}
\end{figure}

Chinese character recognition can be treated as a fine-grained visual classification (FGVC) task. Conventional fine-grained classification task is among the most challenging tasks in computer vision due to subtle inter-class differences and large intra-class variations \cite{wei2021fine}, and has received significant attention over the past decades. However, as illustrated in Figure \ref{fig:CLGF}, unlike birds or other natural objects  \cite{CUB}, resembling Chinese characters present a unique challenge in fine-grained classification, since there are no common semantic parts among the characters. This inherent variability and lack of explicit semantic meanings in glyphs add further challenges to standard FGVC approaches that mainly rely on locating consistent key parts in the objects. Moreover, as can be observed from Figure \ref{rcc-example},  many Chinese characters have resembling glyphs, with very subtle difference, making it an extremely fine-grained visual classification task that is challenging than ever before. 

Today, scene text recognition (STR) in the wild plays a crucial role in urban scene understanding, digital maps and smart cities \cite{OCRSurvey, zheng2024cdistnet}, which deals with natural scene text images captured in the wild via smart phones or cameras, and the challenges include complex backgrounds, various fonts, noise, calibration issues, and imperfect imaging conditions \cite{yu2021benchmarking}.

In this work, we investigate the problem of extremely fine-grained visual recognition over resembling glyphs in the wild. We first build two fine-grained recognition datasets that focuses on resembling glyphs captured under low-data regimes in natural scenes, next devise a novel contrastive learning approach, CCFG-Net, for effectively classifying images having resembling glyphs/characters taken in natural scenes. Indeed, the extremely fine-grained recognition challenges posed by the structure and variability of the Chinese characters and English letters necessitate a more effective approach than traditional fine-grained classification methods. Our method aims to overcome these challenges by focusing on the discriminative features that are critical for effective character classification.

The main contributions of this work are as follows:

\begin{itemize}
\item We build benchmark datasets for extremely fine-grained recognition over resembling glyphs in the wild, which are RCC-FGV for resembling Chinese characters discrimination, and EL-FGVC for resembling English letters identification. 

\item We demonstrate that recognizing Chinese characters in the wild is an intermediate task between classical categorization and fine-grained classification.

\item We devise a deep architecture for this task, incorporating classification and contrastive learning in both Euclidean and angular spaces, in which contrastive learning is applied at the supervised (one against many) and pairwise (one versus one) levels.

\item We evaluate our approach on the RCC-FGVC and EL-FGVC datasets using five different encoders/backbones, and  provide comparisons with representative fine-grained classification methods, which demonstrate the effectiveness of our proposed approach in the extremely fine-grained recognition task of discriminating resembling glyphs in the wild. The introduced datasets and the proposed approach are available at our GitHub repository at \href{https://github.com/faresbougourzi/CCFG-Net}{CCFG-Net}.
\end{itemize}

%

The remainder of the paper is organized as follows: the related works are presented in Section \ref{rtworks}. our benchmark dataset and proposed approach are depicted in Sections \ref{dataset} and \ref{prapp}, respectively. Experiments and results are summarized in Section \ref{exprs}. Sections \ref{abs} and \ref{discu} are devoted to ablation studies and results discussion. The paper is concluded in Section \ref{conc}.

\section{Related Work}
\label{rtworks}
In this section, we discuss the main related works to our approach, focusing on contrastive learning and fine-grained learning. Our approach is primarily based on contrastive learning, while resembling Chinese characters can be considered a fine-grained classification task. Thus, fine-grained approaches are key comparison points in our experiments.

\subsection{Contrastive Learning}

Metric learning from original data for recognition tasks is crucial for applications like face recognition, image retrieval, face verification, zero-shot learning, and person re-identification. With the advent of deep neural networks in computer vision, there has been a shift from shallow models to deep encoders \cite{pmlr-v162-deng22c, bao2022rethinking, pmlr-v139-roth21a, zhu2021visual, Gonzalez-Zapata_2022_CVPR}. The main goal is to develop deep models capable of producing discriminative features.

Most deep metric learning approaches rely on pattern pairs. Formally, encoders in these architectures are trained with loss functions based on pairwise similarities or dissimilarities in the projection space. Representative methods include contrastive loss \cite{Hadsell2006}, triplet loss \cite{Hoffer2015}, N-pair loss \cite{Sohn2016}, Include and Exclude (IE) loss \cite{WU2018}, angular loss \cite{wang2018additive}, and methods using negative examples \cite{Robinson2021}. These methods require inputs to be prepared as pairs, triplets, or quadruplets during training.

Recently, self-supervised learning has emerged as a tool for model learning using pretext tasks not directly related to the main task \cite{Cheng2021, Chen2021, Henaff2020, zbontar2021barlow, Moon2022}. Some self-supervised methods are non-contrastive, employing auxiliary handcrafted tasks to learn representations. Recent contrastive methods have leveraged self-supervision \cite{Kim_2022_CVPR} and data augmentation \cite{Chen2020}. For instance, in \cite{Chen2020}, the authors propose SimCLR, an unsupervised framework for contrastive learning of visual representations. SimCLR is based on two main concepts: generating different images (positive pairs) from the same original image through data augmentation and using a learnable nonlinear transformation between the representation and the contrastive loss.

In \cite{khosla2020supervised}, the idea of SimCLR \cite{Chen2020} is extended to the supervised case, aiming to learn a deep feature space where samples of the same class are moved closer together while samples from different classes are moved apart. The training of supervised contrastive learning (SCL) consists of two stages. In the first stage, the projection head and the backbone encoder are trained together to perform contrastive metric learning based on the projection of the deep representation of two copies of a given batch of labeled images. The loss function forces images of the same class to be close to each other in the projection space while separating images of different classes. In the second stage, the projection head is removed, and a new classification head is attached to the encoder, which is frozen, meaning only the classification layer is learned.

In \cite{Grill2020}, the authors introduce BYOL, a self-supervised method for learning representations involving a deep encoder that is iteratively bootstrapped. BYOL is more robust to the choice of image augmentation type than contrastive methods because it does not use negative pairs. BYOL uses two neural networks, called Online and Target, which learn from each other.




\subsection{Fine-grained Visual Recognition}

Visual recognition tasks can be broadly classified into three categories: (i) basic-level category classification (e.g., generic images \cite{deng_imagenet_2009}), (ii) fine-grained classification (e.g., bird species \cite{CUB, sun2018multi} or fruits \cite{hou2017vegfru}), and (iii) instance-level analysis (e.g., person \cite{ye2021deep} or car re-identification \cite{bai2021disentangled}). Fine-grained recognition tasks are particularly challenging due to subtle inter-class differences and large intra-class variations \cite{wei2021fine}. Over the last decade, significant progress has been made in fine-grained recognition, driven by advances in deep learning techniques. Various approaches for fine-grained recognition have been proposed, which can be classified into the following categories: part-based methods, attribute-based methods, metric learning methods, attention-based methods, and multimodal methods.

Convolutional Neural Networks (CNNs) have been extensively used for fine-grained recognition tasks \cite{zhuang2020learning, zhang2021progressive, liang2022penalizing, rao2021counterfactual}. Inspired by the human cognitive system, P. Zhuang et al. proposed the Attentive Pairwise Interaction Network (API-Net) \cite{zhuang2020learning}, which compares fine-grained pairs to identify contrastive clues. API-Net consists of: (i) mutual vector learning, (ii) gate vector generation, and (iii) pairwise interaction. Initially, a backbone CNN is used to obtain deep features of pairs, which summarize the contrastive cues. These cues are compared with the extracted deep features to create gate vectors, detecting distinguishable clues for the pairs. Finally, residual attention blocks perform pairwise interaction between the gate vectors and the deep features.

Y. Liang et al. introduced Moderate Hard Example Modulation (MHEM) to address overfitting on difficult examples during training, which leads to misclassification of challenging examples in testing \cite{liang2022penalizing}. MHEM employs three conditions—hard mining, moderate, and moderately sensitive—to appropriately scale the losses associated with hard examples, effectively preventing overfitting and enhancing the model's ability to classify difficult instances.

Recently, Transformers have shown remarkable capabilities in fine-grained tasks, as demonstrated by works such as He et al.'s \cite{he2022transfg} and Sun et al.'s \cite{sun2022sim}. In \cite{he2022transfg}, Ju. He et al. introduced the Transformer-based Part Selection Module, designed to identify the most distinctive patches from the input image before the final layer of the transformer. To enhance discerning power between fine-grained classes in the decision space, they employed contrastive loss. In \cite{sun2022sim}, H. Sun et al. addressed limitations of previous part-based approaches that focused solely on discriminative regions while neglecting the overall object structure. Their Structure Information Modeling Transformer (SIM-Trans) approach consists of: (i) structure information learning (SIL) and (ii) multi-level feature boosting (MFB). The SIL module detects discriminative regions while preserving the structural composition of objects, considering the entire object structure. The MFB module leverages multi-level features and uses contrastive learning to handle significant similarities between foreground classes effectively.


\section{The Resembling Glyph Datasets}
\label{dataset}

We build two resembling glyph datasets, which are the Chinese resembling character dataset and the English resembling letters dataset. Among them, the construction of the resembling Chinese character dataset (RCC-FGVC) involves two stages, which are the building of the resembling Chinese character dictionary (RCD) and the collection and annotation of resembling Chinese characters based on RCD. The resembling English letters dataset will be introduced in the meantime. 

\subsection{Building the Chinese Resembling Character Dictionary}

Due to the unavailability of a comprehensive resembling Chinese character dictionary, the first step in constructing resembling Chinese character dictionary (RCD)  is to identify resembling groups of Chinese characters. Three strategies have been used for this purpose.

The first strategy involves summarizing the available dictionaries of resembling Chinese characters. This involved searching bookstores. In total, we found six dictionaries or booklets that introduce groups of resembling Chinese characters \cite{dict11, dict12}. With the aid of a research assistant, we  manually summarized the groups of resembling Chinese characters in these dictionaries/booklets, in which we obtained 755 groups of resembling Chinese characters. Most groups contain two or three resembling characters, while others contain four or five characters.

The second strategy involves a trial-and-error method. We examined and analyzed the prediction logs of the ASTER scene text recognition algorithm \cite{aster} on the CTW and RCTW datasets \cite{shopsign}. We used ASTER (a typical sequence learning-based scene text recognition approach) predictions to automatically identify the wrongly predicted characters within sequences. This method yielded pairs of ground-truth characters and their possible resembling counterparts (the wrongly predicted characters). These pairs were then analyzed, and by focusing on the most frequently occurring wrongly predicted character pairs, a research assistant verified whether the two characters resembled each other based on the wrong prediction rate and glyph similarities. This approach resulted in 1343 pairs of Chinese resembling characters. Since the CTW and RCTW datasets contain English letters as well, the same strategy was used to build the English resembling letters dictionary (ERD).

In the third strategy, we used the Campana-Keogh (CK-1) video compression-based distance measure \cite{ckdistance} to exhaustively compute the similarity of every possible Chinese character pair. First, font template images of each Chinese character with a transparent background were generated. Then, CK-1 was used to identify the resembling pairs from the font template images. Due to the computational demands of CK-1 and the fact that there are more than 40 million Chinese character pairs, it took more than one month to complete the computation. Finally, we set an empirical similarity threshold and manually checked the remaining pairs of characters to verify whether they resembled each other. In the end, we obtained 3547 pairs of Chinese resembling characters. It is worth noting that, to the best of our knowledge, this approach is among the earliest attempts to employ computational methods to exhaustively identify resembling Chinese character groups.

In the final stage, the three  dictionaries of resembling Chinese characters obtained above were cross-checked to create the final RCD. Our analysis revealed that the three dictionaries contained complementary and overlapping resembling character groups. During the merging of the three sub-dictionaries, two domain experts verified the new resembling dictionary, resulting in a total of 4366 groups of resembling Chinese characters. In the final dictionary, 4178 groups consist of pairs of resembling characters, 173 groups consist of triplets of resembling characters, 11 groups consist of quadruplets of resembling characters, 3 groups contain five resembling characters, and 1 group contains seven resembling characters. This comprehensive approach ensured a thorough identification and categorization of resembling Chinese characters, facilitating further research and applications in character recognition and related fields.

\subsection{Data Collection}

To build the  Resembling Chinese Character (RCC-FGVC) dataset based on the constructed resembling character dictionary (RCD), we utilize three natural scene text datasets containing Chinese characters: CTW \cite{ctw}, RCTW \cite{rctw}, and ReCTS \cite{rects}. These datasets inherently present a wide range of challenges as they were captured in the wild. Both CTW \cite{ctw} and ReCTS \cite{rects} datasets include sequence and character-level annotations. The RCTW \cite{rctw} dataset, however, only provides sequence-level (instance-level) annotations. For CTW and ReCTS, we directly used the character annotations as ground truth. For the RCTW dataset, we manually annotated the characters at the character level. Finally, using the character-level annotations from these three datasets, we cropped the scene character images for each Chinese character that appears in the resembling dictionary and organized them by character classes/categories.

The obtained dataset has varying numbers of samples per class, with some characters having an excessively large number of samples and many others having few samples. This variation is due to the nature of the datasets as natural scene images in the wild (e.g., street-view images), in which most of the Chinese characters come from signboards. Consequently, many characters have few samples since they are rarely used in signboards, while other characters are widely used, such as those for ``electricity''  and``shops''. To build a low-shot Chinese resembling character dataset, we filter out characters with less than 20 samples, then randomly select 20 images for characters with more than 20 samples. Finally, the resulting low-shot Chinese resembling character dataset contains 624 Chinese characters, with each character represented by 20 samples. The samples of each character is randomly split into 8:4:8 samples for training, validation and testing, respectively.

To build a low-shot resembling English letter (EL-FGVC) dataset, we use the ICDAR 2013 robust reading competition dataset \cite{icdar2013}, which includes character-level annotations. For each class (letter/number) in the English resembling letters dictionary, we crop the corresponding images from the ICDAR 2013 dataset. Similar to the process for the Chinese resembling character dataset, we filter out classes (i.e., letters/numbers) with inadequate samples and then artificially balance the number of samples in the remaining classes. Finally, the resulting English resembling letter dataset contains 43 classes (letters/numbers), with each class having 20 samples.

\subsection{Dataset Characteristics and Challenges}
\label{rchall}

Figure \ref{rcc-example} and Table \ref{tab:rel} depict some Chinese resembling character pairs and resembling letters/numbers in the English resembling letter dictionary, respectively. These examples highlight characters that have subtle differences and are extremely challenging to distinguish.

\begin{table}[H]
 \caption{Examples of English Resembling letters/numbers.}
 \begin{center}
\label{tab:rel}
\centering
\begin{tabular}{ |cc||cc||cc||cc||cc| }
 \hline
1& I&  5& S&  2& Z&  4& A & 9& g \\
 \hline
8& B&  6& G&  0& O&  M& W & m& w\\
 \hline
g& p&  t& f&  l& i&  L& I & F& E\\
 \hline
G& C&  Q& O&  V& Y&  J& I & M& N\\
 \hline
\end{tabular}
\end{center}
\end{table}



In summary, our task is essentially an extremely fine-grained visual classification (FGVC) challenge with the following characteristics: i) dealing with resembling objects/characters with extremely subtle differences, as many Chinese characters look very similar, making the recognition very challenging; ii) small sample learning, as our dataset contains only 20 samples per class, posing a significant challenge to deep learning based approaches; iii) characters, unlike objects in other fine-grained classification tasks such as birds or vehicles, lack common or explicit semantic parts, making their recognition  more challenging  than conventional FGVC tasks; and iv) scene characters in the wild, often  suffer from complex backgrounds, various fonts, noise,  imperfect imaging conditions, and irregular text deformations, making them vary greatly in imagery. Putting the above challenges together, classifying and discriminating low-shot characters/letters with extremely subtle differences becomes an  a new and extremely challenging fine-grained visual classification task that deserves research attention from the community. 


\section{Methodology}
\label{prapp}

\begin{figure*}
\centering
\includegraphics[width=15cm, height=10cm]{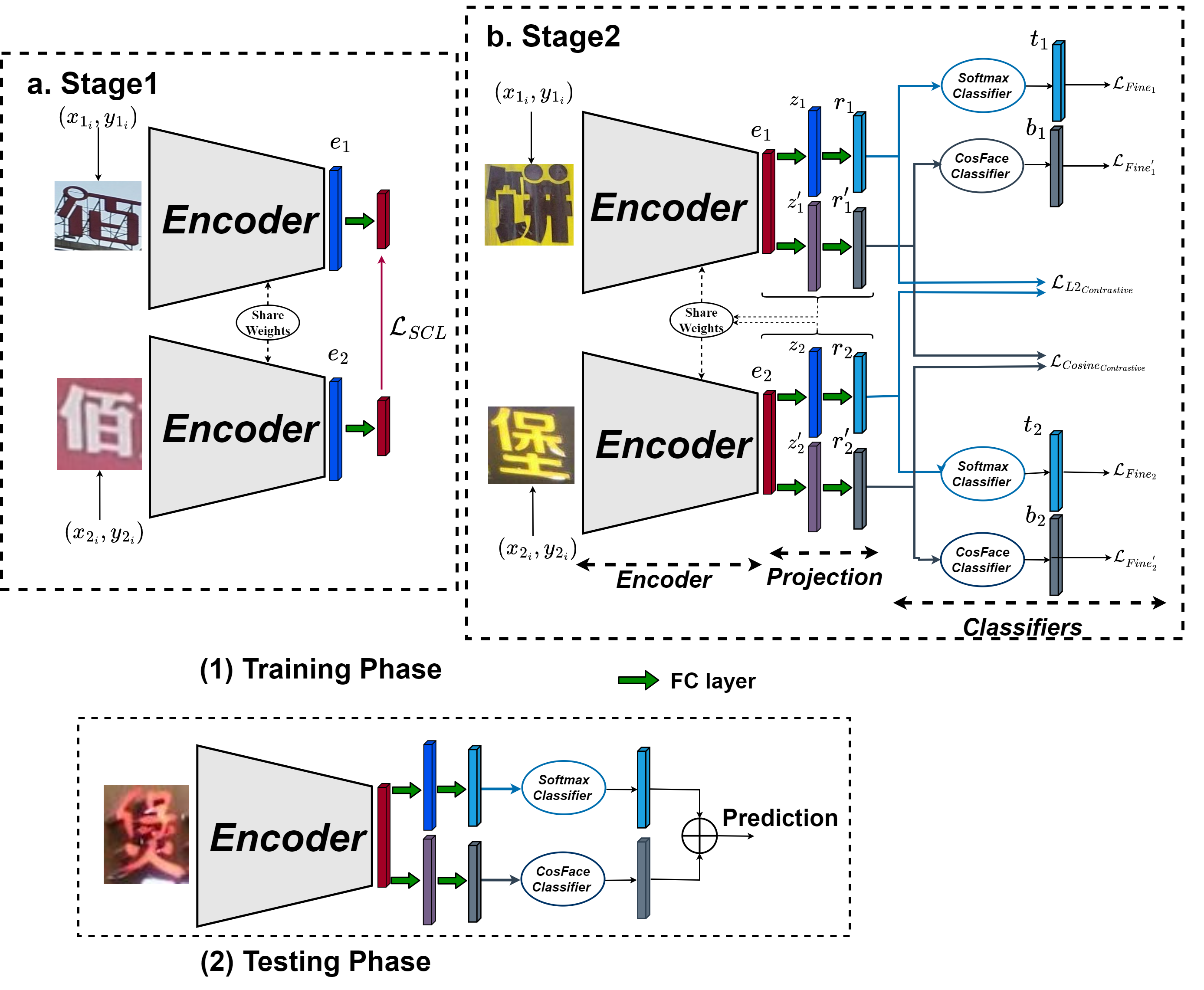}
\caption{An overview of the proposed approach. In which, two training stages are proposed and a testing phase is depicted. } 
\label{fig:CLGF2}
\end{figure*}



Our primary objective is to classify images of resembling glyphs taken from outdoor scenes. As discussed in Section \ref{rchall}, this task is extremely challenging. A trivial solution is to train a pretrained deep learning model to perform the classification. However, due to the presence of a large number of resembling classes and low-resolution images, this approach is unlikely to yield accurate results. Classical deep learning architectures struggle to provide a well-discriminated space for deep class representation, such limitations will be demonstrated in the experimental section. To tackle the problem, we propose a deep neural network that jointly learns to classify classes and perform metric  learning in both Euclidean and Angular spaces for deep feature representation and classification.

\subsection{Deep architecture}

The proposed neural network architecture is shown in Figure \ref{fig:CLGF2}. It  adopts a Siamese like multi-task learning architecture, where two neural networks (shared parameters) to be learned.  In this architecture, each neural network is composed of three modules (see Figure \ref{fig:CLGF2}.b): (i) an encoder $e_{\theta}$, (ii) two parallel branches handling dual spaces Euclidean and angular. Thus, the first branch is  formed  by a projection head $p_{\phi_e}$ followed by a classification head $c_{w_{e}}$. The second branch is formed by a projection head $p_{\phi_a}$ followed by a classification head $c_{w_{a}}$. The first classification head $c_{w_{e}}$ generates the class probabilities that feed  the classic Softmax loss.  The second head $c_{w_{a}}$  provides the class probabilities that feed the large margin cosine loss (LMCL). At inference time, only one architecture is used. At inference time, the class probabilities provided by the two classification heads are averaged.

The aim of the training is to learn the weights of the encoder and projection modules and the classification heads, namely $\theta$,  $\phi_e$, $\phi_a$, $w_e$ and $w_a$. Our training comprises two stages: (i) supervised contrastive learning, (ii) joint pairwise contrastive learning and classification in dual spaces.

The goal of the first stage is to properly initialize the encoder and projection modules so that the class representation (provided by the projection module) can be discriminative. The goal of the second stage is to learn and fine-tune the entire architecture by promoting simultaneous pairwise contrastive learning in Euclidean space and angular space, and classification in these two spaces.

\subsection{First stage of training}

The first stage of training employs a contrastive learning scheme similar to that described in \cite{khosla2020supervised}. For each input image $\mathbf{x}_i$ in the training batch $B$, the encoder and projection modules generate the deep features denoted by $\mathbf{z}_i$, which are then $\ell_2$ normalized, i.e., $\mathbf{z}_i$ is a unit vector. In this stage, the encoder and the projection head weights $\theta$ and $\phi$ are trained by minimizing the following contrastive loss function:

\begin{equation}
    {\cal {L}}_{SCL} = \sum_{i \in B} \frac{-1}{| P (i)| }  \sum_{p \in P (i) }  \log \,  \frac {\exp (\zvect_i  .  \zvect_p / \tau)}  {\sum_{j \in {B}; j \neq i}  \exp (\zvect_i  .  \zvect_j / \tau)   }
    \label{eq:SCLloss}
\end{equation}

Here, the $``\cdot"$ symbol denotes the inner (dot) product,  $\tau$  is a positive scalar temperature parameter, $P(i) =\{  p \in {B} \setminus {i}:   {y}_p = {y}_i \}$ is the set of indices of all positive examples in the current batch $B$
distinct from $i$, and $ |P(i)|$ is its cardinality.  $y_i$ denotes the class label of image $i$.  Unlike \cite{khosla2020supervised} which uses two views of the same image, we adopt a classic  batch that contains a subset of images. We emphasize that contrastive learning in the first stage of training aims at learning the class concept in a deep feature representation. On the other hand, contrastive learning in the second stage aims at reinforcing the similarity of positive pairs and the dissimilarity of negative pairs.

\subsection{Second stage of training}

The aim of the second training stage is to fine-tune the weights of the encoder and projection modules $\theta$,  $\phi_e$,  and $\phi_a$, and to learn the weights of the appended classification heads,  $w_e$ and $w_a$ (Figure \ref{fig:CLGF2}, Stage 2). In this case, the training batch is composed of $N$ image pairs $(\mathbf{x}_{1_j}, y_{1_j})$, $(\mathbf{x}_{2_j}, y_{2_j})$.  $y_{1_j}$ and $y_{2_j}$ denote the class labels of $\mathbf{x}_{1_j}$ and $\mathbf{x}_{2_j}$, respectively. The two images $\mathbf{x}_{1_j}$ and $\mathbf{x}_{2_j}$ in the $j$th pair are either in the same class (positive pair) or in different classes (negative pair).

\begin{figure}
\centering
\includegraphics[width=9cm]{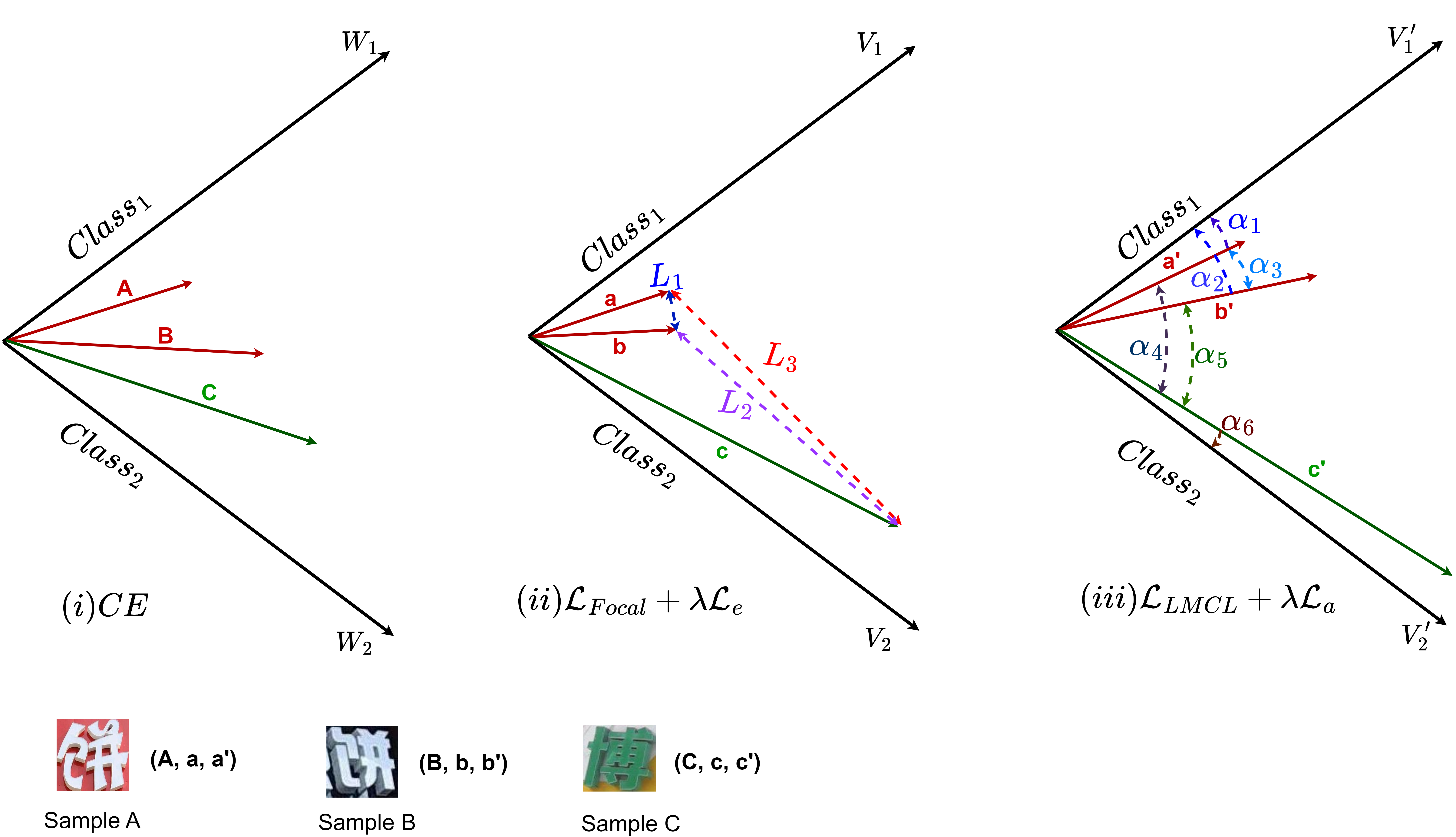}
\caption{ \begin{CJK*}{UTF8}{gbsn}The expected  effect of losses on the deep features of three images. We consider three deep spaces: (i)  the standard CE loss function, (ii) The first head of our CCFG-Net approach (${\mathcal {L}}_{Focal} +  \lambda \,  {\mathcal {L}}_{e}$), and (iii) The second head of our CCFG-Net approach (${\mathcal {L}}_{LMCL} +  \lambda \,  {\mathcal {L}}_{a}$). The labels of Samples A and B is $class_1$ (饼) and the label of Sample C is $class_2$ (博). Sample A deep features in CE, z and z' are  A, a and a' spaces, respectively.  Sample B deep features in CE, z and z' spaces are  B, b and b', respectively. Sample C deep features in CE, z and z' spaces are  C, c and c', respectively. The objective of the first head is to minimize distance $L_1$ and maximize the distances  $L_2$ and $L_3$. On the other hand, the objective of the second head is to  minimize $\alpha_1$ $\alpha_2$ $\alpha_3$ and  $\alpha_6$ and maximize $\alpha_4$ and $\alpha_5$. \end{CJK*}}
\label{fig:lossRes}
\end{figure}

Let $\mathbf{x}_1$ and $\mathbf{x}_2$ be two images corresponding to a given pair. According to the figure, the deep features of the first image (obtained by the encoder and the projection head), $\mathbf{z}_1$, are fed to two classification heads. First, $\mathbf{z}_1$ is projected to two different vectors $\mathbf{r}_1$ and $\mathbf{r}_1^{\prime} \in \mathbb{R}^{C}$ ($C$ denotes the number of classes). These are converted to two probabilities vectors $\mathbf{t}_1$ and $\mathbf{b}_1 \in \mathbb{R}^{C}$ using the classic softmax and the LMCL softmax. Similarly, the second image $\mathbf{x}_2$ will provide the two probabilities vectors $\mathbf{t}_2$ and $\mathbf{b}_2 \in \mathbb{R}^{C}$ (see Figure \ref{fig:lossRes}).

At the level of visual recognition, the task of recognizing resembling glyphs can be viewed as an intermediate step between classical categorization and fine-grained classification. Since classification losses are best suited for general categorization problems and deep metric learning is suitable for purely fine-grained problems, we will rely on hybrid losses encompassing classification losses and metric learning losses.

\textbf{Classification losses:} The classification loss for the two probabilities vectors $\mathbf{t}_1$ and $\mathbf{t}_2$ is given by the classic focal loss:

\begin{align}
\mathcal{L}_{\text{Focal}} = & - \sum_{j=1}^{N} \left[ (1-t_{1}(y_{1j}) )^\gamma \, \log(t_{1} (y_{1j})) \right. \nonumber \\
& \left. + \, (1-{t_2}(y_{2j}))^\gamma \, \log(t_{2} (y_{2j})) \right]
\end{align}

where $t_1 (l)$ is the $ l-$th element of the vector $\tvect_1$ and  $y_{1j}$ is the ground-truth label of image ${\xvect_1}_{j}$ (first image in the $j$-th pair). $\gamma \in [0, 3]$ is the hyperparameter of the focal loss function. 

The classification loss for the two probabilities vectors $\mathbf{b}_1$ and $\mathbf{b}_2$ is given by the LMCL loss:

\begin{equation}
\begin{split}
\label{classifCOSFACE}
&{\mathcal {L}}_{LMCL}= - \sum_{j=1}^N  \\
& \log \frac{\exp \left(s \cdot \left(\cos \theta_{y_{1j}}-m_c \right) \right)}{\exp\left(s \cdot\left(\cos \theta_{y_{1j}}-m_c \right) \right)+ \sum_{k=1, k \neq y_{1j}}^C \exp\left(s \cdot \cos \theta_k) \right)} \\
& +   \log \frac{\exp \left(s \cdot \left(\cos \theta_{y_{2j}}-m_c \right) \right)}{\exp\left(s \cdot\left(\cos \theta_{y_{2j}}-m_c \right) \right)+ \sum_{k=1, k \neq y_{2j}}^C \exp\left(s \cdot \cos \theta_k) \right)} 
\end{split}
\end{equation}

where $m_c \in [0, 0.5]$ is a cosine margin and $s$ is a scaling factor for preventing too small gradients during the training. $\theta_{y_{1j}}$ denotes the angle between the deep representation vector $\mathbf{r}^{\prime}_{1j}$ of the image $\mathbf{x}_{1j}$ and the center of the ground-truth class $y_{1j}$ (i.e., the $y_{1j}$-th row of the classification projection matrix in the head $c_{wa}$).

\textbf{Pairwise contrastive losses:} We propose the following contrastive loss $\mathcal{L}_{e}$ that promotes pairwise contrastive learning in Euclidean space. This loss is given by:

\begin{equation}
{\mathcal {L}}_{e} = \sum_{j=1}^{N}  f_j \;    {dist}^2 ( {\zvect_1}_j, {\zvect_2}_j) +  ( 1 -f_j) \;    \{ max [ 0, m_e - dist (  {\zvect_1}_j, {\zvect_2}_j) ] \}^2  
\label{eq:contrastive-Euclidean}
\end{equation}

where $f_j$ is a binary flag defined as follows: $f_j = 1$ if the pair $j$ is positive, $f_j = 0$ if the pair is negative. $m_e$ denotes the margin, set to one in our implementation.

We also propose the following contrastive loss $\mathcal{L}_{a}$ that promotes pairwise contrastive learning in angular space. The loss is given by:

\begin{equation}
{\mathcal {L}}_{a} = \sum_{j=1}^{N}  f_j \; dist^2 ( \hat{\rvect_1}_j, \hat{\rvect_2}_j) +  ( 1 -f_j) \;    max [ 0, m_a - dist (  \hat{\rvect_1}_j, \hat{\rvect_2}_j) ]^2 
\label{eq:contrastive-angular}
\end{equation}

where $\hat{\mathbf{r}}_1$ denotes the unit vector associated with the vector $\mathbf{r}^{\prime}_1$. $m_a$ is the angular margin. Since the vectors involved in the above distances are unit vectors, it follows that the above loss minimizes the angle between the deep features $\mathbf{r}^{\prime}_1$ and $\mathbf{r}^{\prime}_2$ for a positive pair and maximizes that angle for a negative pair. Indeed, the above loss can be expressed as:

\begin{equation}
\begin{split}
{\mathcal {L}}_{a} = & \sum_{j=1}^{N}  f_j \;  2 \, ( 1 - \cos (\alpha_j) )  \\
& +  ( 1 -f_j) \;    \{ max [ 0, m_a - \sqrt{   2 \, ( 1 - \cos (\alpha_j)) }    ] \}^2
\label{eq:contrastive-cosine}
\end{split}
\end{equation}

where $\alpha_j$ is the angle  between the deep features ${{\rvect}_1}_j$ and ${{\rvect}_2}_j$.

The global loss used in end-to-end training of the proposed network is given by:

\begin{equation}
\label{totalloss}
{\mathcal {L}} = {\mathcal {L}}_{Focal} +  {\mathcal {L}}_{LMCL} + \lambda \, ( {\mathcal {L}}_{e} +  {\mathcal {L}}_{a} )
\end{equation}

where $\lambda$ is a balance parameter.

In each epoch, all possible positive pairs are chosen, while for the negative pairs, one image is selected randomly from a set of $p_n$ random classes for each image in the training dataset. The purpose of selecting $p_n$ negative pairs is to balance the number of negative and positive pairs. Additionally, the selection of different negative pairs each epoch helps prevent overfitting and promotes the development of a discriminative space between all classes.

\begin{table}[ht!]
 \caption{Training Settings. Opt, Lr Sc, Bs, $D z_1$ and N epochs are the used optimizer, learning rate schedule, Batch size, dimension of $z_1$ and the number of training epochs, respectively. MSLS and WCLS are multi-steps learning schedule and Worm-Cosine learning schedule, respectively.}
 \begin{center}
\label{tab:trdt}
\centering
\scalebox{0.5}{\begin{tabular}{|l|l|c|c|c|c|c|c||c|c|c|c|c|c|}

\hline
  {\multirow{3}{*}    \textbf{Backbone}}   & {\multirow{3}{*}    \textbf{Tr Stage}}  & \multicolumn{6}{|c|}{\textbf{RCC-FGVC
}}& \multicolumn{6}{|c|}{\textbf{REL-FGVC}} \\
\cline{3-14}


 &   & \textbf{Opt} &\textbf{Lr Sc}&\textbf{Ini Lr} &   \textbf{Bs}& \textbf{$D z_1$ }  & \textbf{N epochs} &\textbf{Opt} &\textbf{Lr Sc} &\textbf{Ini Lr}  &   \textbf{Bs}& \textbf{$D z_1$ }  & \textbf{N epochs}
 \\
\hline

ResNet-50&  1&  Adam  &  MSLS & $10^{-4}$ &  128   &  -  & 100 &Adam  &  MSLS  & $10^{-4}$ &64   &  -  & 100
\\\cline{2-14} 

&  2 & Adam   &  MSLS  & $10^{-4}$ &   128  & 1800  & 25 & Adam   &  MSLS  & $10^{-4}$ &   64  & 512  & 25

\\\hline\hline

Densenet-&  1 & Adam   &  MSLS  & $10^{-4}$ &  32   &   -  &100  & Adam   &  MSLS& $10^{-4}$   &  32   &  -  &100
\\\cline{2-14}

161&  2& Adam   &  MSLS& $10^{-4}$ &   32  &   1800  & 25 &   Adam   &  MSLS &  $10^{-4}$ &   32  &   512  & 25

\\\hline\hline

EfficientNet&1 & Adam  &  MSLS &  $10^{-4}$ &  32   &   -  & 100 & Adam  &  MSLS &  $10^{-4}$ &  32   &   -  & 100  \\\cline{2-14}

-B5 &  2&  Adam  & MSLS &  $10^{-4}$  &   32  &  1800   & 25 & Adam  & MSLS  &  $10^{-4}$ &   32  &  512   & 25

\\\hline\hline


ViT (Base)& 1& SGD  &  WCLS &  $10^{-3}$ &  32 &  -  & 100 & SGD  &  WCLS  &  $10^{-3}$ &  32 &  -  & 100
\\\cline{2-14}

& 2&  SGD  &  WCLS &  $10^{-4}$ &  32   & 512  & 25 &  SGD  &  WCLS  &  $10^{-4}$ &  32   & 512 & 25
\\\hline\hline
ViT (Large)& 1& SGD   &  WCLS &  $10^{-3}$ &  32  &  - & 100 & SGD   &  WCLS  &  $10^{-3}$ &  16  &  - & 100
  \\\cline{2-14}

&2 & SGD   &  WCLS &  $10^{-4}$  &  16   &  1024 & 25 & SGD   &  WCLS  &  $10^{-4}$ &  16   &  512 & 25

\\\hline
\end{tabular}}
\end{center}
\end{table}


\section{Experiments and Results}
\label{exprs}
\subsection{Experimental Settings}

In order to assess the effectiveness of our approach, we conduct evaluations using both CNN and Transformer backbones as encoders. We aim to demonstrate the efficiency of our approach regardless of the choice of backbone encoder. The following backbones are considered for evaluation purposes: ResNet-50, DenseNet-161, EfficientNet-B5, ViT Base, and ViT Large. These backbone architectures are selected based on their established performance and widespread usage in various computer vision tasks. By evaluating multiple architectures, we aim to analyze the impact of different backbone designs on the overall performance of our approach. The experiments are done using Pytroch and timm libraries with NVIDIA GPU Device GeForce RTX 3090 24 GB.

Since the evaluated datasets and backbones have different sizes and characteristics, different training hyperparameters were adopted. Table \ref{tab:trdt} summarizes the training hyperparameters used for each dataset, backbone, and training stage. For our approach, a multi-step learning rate decay schedule is utilized for CNN backbones, while a warm cosine learning schedule is employed for Transformer backbones. The following data augmentations are applied: Color Jitter, Random grayscale transformation, and Random Rotation.

To train our approach in the second stage, all possible positive pairs are constructed. On the other hand, for the negative pairs ($p_n$), we set $p_n$ to four and six for RCC-FGVC and EL-FGVC datasets, respectively. This values balances between the number of positive and negative pairs, and it proves the right trad-off between the performance and training time. SCL loss function hyper-parameters (temperature and base\_temperature) are set to 0.07. Focal loss function hyper-parameter gamma is set to 3.5. CosFace loss function hyper-parameters s and m are set to 30.0 and 0.40, respectively. Both contrastive losses margin hyper-parameter is set to 1.0, and the balancing hyper-parameter $\lambda$ is set to 0.3.

\subsection{Resembling Chinese Characters Recognition}

To evaluate the performance of our approach for Resembling Chinese Characters Recognition (RCCR), we investigated both CNN and Transformer backbones, including ResNet-50 \cite{he2016deep}, DenseNet-161 \cite{huang2017densely}, EfficientNet-B5 \cite{tan2019efficientnet}, and ViT architectures (Base and Large) \cite{dosovitskiy2020image}. Training these backbones with a decision layer corresponding to the number of classes and using the CE loss function served as the baseline approach. As shown in Table \ref{tab:mainrCC}, our approach surpasses the baseline by a considerable margin, with test accuracies improved by 30.11\%, 21.13\%, 16.09\%, 25.64\%, and 24.23\% for ResNet, DenseNet, EfficientNet, ViT Base, and ViT Large backbones, respectively. In terms of F1-score, our approach improved performance by 28.36\%, 21.44\%, 17.46\%, 26.08\%, and 24.44\%, respectively. Additionally, the results indicate that transformer backbones outperform CNN-based ones, with ViT Large achieving the best performance and ResNet-50 ranking fifth.

To compare the performance of the proposed approach with the state-of-the-art approaches, we conducted extensive experiments using various approaches with a consistent backbone architecture. Each approach utilized the same backbone architecture, ensuring a fair comparison between different methodologies.

\begin{itemize}
    \item \textbf{Sophisticated Loss Function:} M2B \cite{liang2022penalizing} introduced a new loss function specifically designed to handle difficult examples.
    \item \textbf{Part-based Methods:} MMAL-Net \cite{zhang2021multi} and TransFG \cite{he2022transfg} focus on leveraging parts or sub-regions of the input for improved performance.
    \item \textbf{Attribute-based Methods:} PCA-Net \cite{zhang2021progressive} and API-Net \cite{zhuang2020learning} utilize attributes or characteristic features of the input data for effective classification.
    \item \textbf{Metric Learning Methods:} CosFace \cite{wang2018cosface}, Softtriple \cite{qian2019softtriple}, and SCL \cite{khosla2020supervised} aim to learn a suitable metric space for better discrimination between different classes.
    \item \textbf{Attention-based Methods:} MMAL-Net \cite{zhang2021multi}, PCA-Net \cite{zhang2021progressive}, CAL \cite{rao2021counterfactual}, and Sim-Trans \cite{sun2022sim} focus on assigning importance or attention to specific regions or features of the input during classification.
\end{itemize}

As depicted in Table \ref{tab:mainrCC}, our approach outperforms all comparison methods for both CNN and Transformer-based approaches on the ECC-FGVC dataset. The leading competitor varies across different backbones, with our approach surpassing ResNet-50's best competitor (SCL \cite{khosla2020supervised}) by 4.43\% in accuracy and 4.34\% in F1-score. Several state-of-the-art methods achieve similar performance (around 72\%), such as Softtriple loss \cite{qian2019softtriple}, API-Net \cite{zhuang2020learning}, M2B \cite{liang2022penalizing}, and CAL \cite{rao2021counterfactual}. However, MMAL-Net \cite{zhang2021multi}, which is part-based, performs worse due to the RCC's lack of unified semantic parts, leading to overfitting. M2B \cite{liang2022penalizing} emerges as a strong competitor, especially for Densenet-161 and EfficientNet-B5 backbones, outperforming metric learning methods. With Densenet-161, M2B exceeds the third-ranked SCL by 7.26\% in accuracy and 6.57\% in F1-score. For EfficientNet-B5, M2B leads by 7.81\% in accuracy and 7.45\% in F1-score over SCL. Despite M2B's strong performance, our approach achieves superior results in terms of F1-score, surpassing M2B by 1.83\% and 4.77\% for Densenet-161 and EfficientNet-B5, respectively. This demonstrates the effectiveness of our approach in enhancing SCL pretrained weights. Notably, our CCFG-Net significantly improves SCL performance, boosting F1-score by 8.4\% and 12.22\% for Densenet-161 and EfficientNet-B5, respectively.

For Transformer based backbones, the metric learning  approaches (Softtriple, SCL  and CosFace) have emerged as the top competitors, surpassing other methods by a significant margin. While other competitors only offer marginal improvements over the baseline results, the metric learning approaches excel in performance. For example, when using the TransFG (ViT Base) approach, the testing data demonstrates a 1.55\% increase in accuracy and a 1.49\% increase in F1-score compared to the baseline. With the Large backbone, the improvements are even more substantial, with a 6.53\% increase in accuracy and a 6.21\% increase in F1-score. Similar trends are observed with Sim-Trans for both Small and Large ViT backbones. Among the metric learning approaches, CosFace ranks as the best competitor for both the Base and Large versions of ViT, closely followed by SCL. Notably, CCFG-Net achieved substantial performance compared with SCL, by a significant margin.  For instance, CCFG-Net (ViT Large) achieves an impressive performance boost of 8.62\% in accuracy and 8.59\% in F1-score on the testing data compared to SCL. Additionally, CCFG-Net (ViT Large) outperforms the leading competitor, CosFace, by 5.77\% in accuracy and 5.88\% in F1-score on the test data. These results highlight the superior performance of CCFG-Net (ViT Large) and establish it as the leading choice among the metric learning approaches for the ViT architecture.

The comparisons conducted demonstrate the effectiveness of the proposed approach for Resembling Chinese Characters Recognition in the wild, particularly in terms of performance when compared to various Fine-Grained approaches. These comparisons highlight the high ability of the proposed approach to effectively handle the challenges posed by resembling characters and the unpredictable conditions encountered in real-world scenarios.

\begin{table}[ht]
 \caption{Performance Evaluation of CCFG-Net Approach and Comparison Methods on RCC-FGVC Dataset with Five Backbones (ResNet, Densenet, EfficientNet, ViT Base, and ViT Large).}
 \begin{center}
\label{tab:mainrCC}
\centering
\scalebox{0.8}{\begin{tabular}{|l|l|c|c|c|c|}

\hline
  {\multirow{2}{*}    \textbf{\centering Backbone}}   & {\multirow{2}{*}    \textbf{\centering Method}}  &\multicolumn{2}{|c|}{\textbf{Validation}}& \multicolumn{2}{|c|}{\textbf{Test}}  \\
\cline{3-6}


 &   & \textbf{Acc} &\textbf{F1-S} &   \textbf{Acc}& \textbf{F1-S}   
 \\
\hline
&  CE loss (Baseline)&  50.28 & 51.50 & 49.09  & 50.94 
\\\cline{2-6} \cline{2-6}

&  MMAL-Net \cite{zhang2021multi}&   21.03 &  19.89 &  20.45  & 20.40   
\\\cline{2-6} 

&  CosFace loss \cite{wang2018cosface}& 57.88  & 56.73 & 58.66  & 58.01
\\\cline{2-6}

& PCA-Net \cite{zhang2021progressive}&  69.83 & 69.28 & 68.86 & 68.73   \\\cline{2-6} 

 & M2B \cite{liang2022penalizing}&  72.08  &  71.34& 70.27 &  69.99  
 \\\cline{2-6}

 Resnet-50& API-Net \cite{zhuang2020learning}& 71.15  & 70.38 &  71.47 & 71.12 \\\cline{2-6}

&  Softtriple loss \cite{qian2019softtriple}&  72.51 & 72.44 &  71.33 & 71.86
 \\\cline{2-6}

& CAL \cite{rao2021counterfactual}& 73.01&72.32 &  72.57  &  72.14 \\\cline{2-6}

&  SCL loss \cite{khosla2020supervised}&  74.82 & 74.56 &  74.77 & 74.96
\\\cline{2-6}

& \bf{CCFG-Net}& \bf{79.89}  & \bf{79.70} &  \bf{79.20}  & \bf{79.30} 
\\\hline\hline

&  CE loss (Baseline)& 62.32  & 61.23 & 61.09  & 60.78 
\\\cline{2-6} \cline{2-6}

& API-Net \cite{zhuang2020learning}& 61.47  &  60.60&  61.57 & 61.18 
  \\\cline{2-6}
  
&  CosFace loss \cite{wang2018cosface}&  68.65 & 67.58 & 67.33  & 66.68 
\\\cline{2-6}

Densenet-161&  Softtriple loss \cite{qian2019softtriple} & 73.58  & 73.58 & 72.81  & 73.23 
\\\cline{2-6} 

&  SCL loss \cite{khosla2020supervised}& 72.59  & 72.43 & 73.26 & 73.82 
\\\cline{2-6}

 & M2B \cite{liang2022penalizing}&  81.28 & 80.29 &  80.52 & 80.39 
  \\\cline{2-6}

& \bf{CCFG-Net}&  \bf{83.68} &  \bf{83.26}
 &  \bf{82.22}  &  \bf{82.22}  
\\\hline\hline

&CE loss (Baseline)& 65.63  & 64.77 & 65.88  & 65.21\\\cline{2-6}

&API-Net \cite{zhuang2020learning} &  55.10 & 55.41 & 5430  & 55.08
   \\\cline{2-6}

& SoftTripple loss \cite{qian2019softtriple} & 68.36  &68.26 & 66.72  &66.95 
  \\\cline{2-6} 
  
EfficientNet-B5 &CosFace \cite{wang2018cosface}& 69.63  & 68.77 & 70.05  & 69.61
\\\cline{2-6} \cline{2-6}

& SCL loss \cite{khosla2020supervised} & 69.79  &69.37  & 70.41  &70.45
\\\cline{2-6}
   
& M2B \cite{liang2022penalizing}&  77.93& 77.52 &  78.22 & 77.90 
  \\\cline{2-6} 

& \bf{CCFG-Net}& \bf{82.58}  & \bf{82.70} & \bf{81.97}  &\bf{82.67} 
\\\hline\hline


& CE loss (Baseline)& 58.18   &56.97  &57.87   & 57.30 
  \\\cline{2-6} \cline{2-6}

 & TransFG \cite{he2022transfg}& 60.44  &59.29 & 59.42   & 58.79
  \\\cline{2-6}
 
& Sim-Trans \cite{sun2022sim}& 59.52  & 6047 & 58.37  & 59.80 \\\cline{2-6}  

ViT (Base)& SoftTripple \cite{qian2019softtriple} &  63.11  & 62.46 & 62.32  &62.11  
  \\\cline{2-6} 
  
& SCL loss \cite{khosla2020supervised}& 76.53  & 75.88 & 74.72  & 74.76 
\\\cline{2-6}

& CosFace \cite{wang2018cosface}&  75.16 & 74.54 & 75.74  & 75.58 
\\\cline{2-6} 

& \bf{CCFG-Net}&  \bf{84.42} &  \bf{84.09} &  \bf{83.51}  & \bf{83.38} 
\\\hline\hline
& CE loss (Baseline)&  61.85 & 61.19& 60.82  &60.58 
  \\\cline{2-6} \cline{2-6}

& Sim-Trans \cite{sun2022sim}&  65.38 & 64.54&65.33   &64.64  \\\cline{2-6}  

ViT (Large)& TransFG  \cite{he2022transfg}& 67.38  & 66.28& 67.35   & 66.79
  \\\cline{2-6}

& SoftTripple  \cite{qian2019softtriple}& 68   & 67.40 & 67.24  &66.94 
  \\\cline{2-6} 

& SCL loss \cite{khosla2020supervised}& 77.01  & 76.61&  76.43 &76.43 
\\\cline{2-6}

& CosFace \cite{wang2018cosface}&  79.94 & 79.54& 79.28  &79.14 
\\\cline{2-6}
 
& \bf{CCFG-Net}& \bf{85.52} & \bf{85.24} & \bf{85.05}  &\bf{85.02}

\\\hline
\end{tabular}}
\end{center}
\end{table}

\subsection{Resembling English Letters Recognition}

Similar to the experiments of Resembling Chinese Characters, the same backbones are evaluated for the Resembling English Letters dataset. Table \ref{tab:mainren} depicts the comparison between our approach and the baseline results, in which the backbone is trained with CE loss function). These results show that our approach boosts the performance of the baseline backbones by considerable margin, where the test set F1-score are enhanced by 28.07\%, 11.61\%, 18.39\%,  19.62\% and 20.83\% for ResNet, Densenet, EfficientNet, ViT Base, and ViT Large backbones, respectively. For our proposed CCFG-Net approach, the comparison between the five evaluated backbones shows that the difference is small unlike the Resembling Chinese Characters dataset. This because there is a big difference in the number of classes and the recognition difficulty between the Chinese characters and English letters. In more details, the worst backbone is ViT Base then Resnet-50, this last outperforms ViT Base backbone by 1.16\% and 1.42\% for the accuracy and F1-score, respectively. On the other hand, the other evaluated backbones achieved very close performance; Densenet-161, EfficientNet-B5 and ViT Large.

The comparison with state-of-the-art approaches are summarized in Table \ref{tab:mainren}, where different types of fine-grained approaches are considered, similar to the RCC-FGVC dataset. For ResNet-50 backbone, it is observed that not only does the MMAL-Net approach \cite{zhang2021multi} perform worse than the baseline (CE loss), but CAL \cite{rao2021counterfactual} and SoftTriple loss \cite{qian2019softtriple} also exhibit similar performance. This highlights the difficulty of recognizing real-world language letters and characters for certain state-of-the-art approaches that were primarily designed for finding similar parts in natural images. On the other hand, the rankings of the remaining state-of-the-art approaches show slight differences compared to the results on the RCC-FGVC dataset. Specifically, PCA-Net \cite{zhang2021progressive} emerges as the top competitor among the comparison approaches, followed by SCL \cite{khosla2020supervised} and M2B  \cite{liang2022penalizing} methods. This variation can be attributed to differences in the difficulty and number of classes between the REL-FGVC and RCC-FGVC datasets. In contrast, our proposed approach, CCFG-Net (ResNet-50), consistently achieves the best performance on both datasets. It outperforms the top competitor by 8.14\% in Accuracy and 8.62\% in F1-score on the testing data.

Among the other two CNN backbones, Densenet-161 and EfficientNet-B5, SCL \cite{khosla2020supervised} demonstrates superior performance compared to other competitors. Following closely behind are the M2B \cite{liang2022penalizing} and CosFace \cite{wang2018cosface} methods, respectively. Notably, only a few of the comparative methods manage to surpass the baseline results, particularly for EfficientNet-B5. In contrast, our approach maintains a significant lead and consistently outperforms all evaluated methods for both backbones.

In the context of Transformer-based backbones, it has been observed once more that the SoftTriple loss underperforms when compared to the baseline. Conversely, the other two metric learning approaches, SCL and CosFace, emerge as strong competitors, with each excelling in its own right. However, our newly proposed CCFG-Net outshines all of the comparison methods, regardless of the language being considered, as each language presents distinct characteristics such as varying numbers of classes and levels of writing complexity.

\begin{table}[ht]
 \caption{Performance Evaluation of CCFG-Net Approach and Comparison Methods on REL-FGVC Dataset with Five Backbones (ResNet, Densenet, EfficientNet, ViT Base, and ViT Large).  }
 \begin{center}
\label{tab:mainren}
\centering
\scalebox{0.8}{\begin{tabular}{|l|l|c|c|c|c|}

\hline
  {\multirow{2}{*}    \textbf{\centering Backbone}}   & {\multirow{2}{*}    \textbf{\centering Method}}  &\multicolumn{2}{|c|}{\textbf{Validation}}& \multicolumn{2}{|c|}{\textbf{Test}}  \\
\cline{3-6}

 &   & \textbf{Acc} &\textbf{F1-S} &   \textbf{Acc}& \textbf{F1-S}  
 \\
\hline
&  CE loss (Baseline)& 66.08&66.45&59.30&58.76
\\\cline{2-6} \cline{2-6}

&  MMAL-Net \cite{zhang2021multi}&    31.97& 30.81  & 34.01  & 33.14
\\\cline{2-6} 

& CAL \cite{rao2021counterfactual}& 52.51 & 50.89 & 56.10 & 55.32\\\cline{2-6}

&  Softtriple loss \cite{qian2019softtriple}&  60.07 & 60.59 & 56.97 & 56.63 \\\cline{2-6}

&  CosFace loss \cite{wang2018cosface}& 65.69 &65.27 &68.31 & 67.59\\\cline{2-6}

Resnet-50 &API-Net \cite{zhuang2020learning}&  69.37 & 68.59 & 71.22 & 71.06 \\\cline{2-6}

 & M2B \cite{liang2022penalizing}& 70.34 & 69.36 & 71.80 & 71.26 \\\cline{2-6}
 
&  SCL loss \cite{khosla2020supervised}& 74.03&73.83 &74.22 &74.57\\\cline{2-6}

& PCA-Net \cite{zhang2021progressive}& 76.93 & 76.68 & 78.77 & 78.21  \\\cline{2-6} 

& \bf{CCFG-Net}&   \bf{85.27} &  \bf{85.35} &  \bf{86.91} &  \bf{86.83}
\\\hline\hline

&  CE loss (Baseline)&  73.83 & 72.07 & 77.03 & 76.57
\\\cline{2-6}

& API-Net \cite{zhuang2020learning}& 65.69 & 64.95 & 69.47 &69.31
  \\\cline{2-6}
  
Densenet-161&  Softtriple loss \cite{qian2019softtriple} & 68.02& 68& 70.63& 70.36
\\\cline{2-6} 
  
&  CosFace loss \cite{wang2018cosface}&   70.93 & 69.60 & 77.90 &77.77
\\\cline{2-6}

& M2B \cite{liang2022penalizing}&  77.51 & 77.57& 78.77&78.63
\\\cline{2-6}
  
&  SCL loss \cite{khosla2020supervised}& 77.90 & 76.50 & 81.10 & 81.62
\\\cline{2-6}

& \bf{CCFG-Net}&    \bf{85.85} &  \bf{86.19} &  \bf{88.37} &  \bf{88.18}
\\\hline\hline

&CE loss (Baseline)&  62.20&  60.64&70.63  & 69.84\\\cline{2-6}

& SoftTripple loss \cite{qian2019softtriple} &  50.58 &  50.25& 58.13 & 58.98
  \\\cline{2-6} 

&API-Net \cite{zhuang2020learning} &   67.44 & 66.65  & 65.69  &  65.16
   \\\cline{2-6}
   
EfficientNet-B5&CosFace \cite{wang2018cosface}&  66.86 & 65.40 & 70.05 &69.79
\\\cline{2-6}

& M2B \cite{liang2022penalizing}& 68.60 & 66.71 & 72.38 & 72.34
  \\\cline{2-6} 
  
& SCL loss \cite{khosla2020supervised} &70.34 & 69.31&75 &75.05
\\\cline{2-6}

& \bf{CCFG-Net}&   \bf{85.85}& \bf{85.74} & \bf{88.37}  & \bf{88.23}
\\\hline\hline


& CE loss (Baseline) &58.72 & 57.14& 65.98& 65.79
  \\\cline{2-6}
  
& SoftTripple \cite{qian2019softtriple}  & 57.55& 58.36 & 56.97 & 57.82 
  \\\cline{2-6}
  
 & Sim-Trans \cite{sun2022sim}&   62.79   & 60.97 & 62.21 & 61.42 \\\cline{2-6} 
 
 ViT (Base) & TransFG \cite{he2022transfg}& 65.50 &  63.08& 70.93 &70.50
  \\\cline{2-6}
  
& CosFace \cite{wang2018cosface}& 75.38 & 74.57 & 77.61 & 77.19
\\\cline{2-6} 

& SCL loss \cite{khosla2020supervised} & 77.71 & 77.11 & 82.55 & 82.15
\\\cline{2-6}

& \bf{CCFG-Net}&   \bf{88.95} & \bf{88.44} & \bf{85.75} & \bf{85.41}
\\\hline\hline
& CE loss (Baseline)&61.04 & 59.43& 67.15 &67.21
  \\\cline{2-6}
  
& SoftTripple  \cite{qian2019softtriple} & 62.20 &  61.80 & 66.27 &  66.35 
  \\\cline{2-6}

& Sim-Trans \cite{sun2022sim}&  71.12 & 70.41 & 74.12  & 73.67 \\\cline{2-6}

ViT (Large)& TransFG  \cite{he2022transfg}&    70.15  &  69.26&   74.41&  74.14
  \\\cline{2-6}
& CosFace \cite{wang2018cosface} &73.64  &  72.18& 78.48 &78.13
\\\cline{2-6}

& SCL loss \cite{khosla2020supervised}&   74.22& 73.51 & 83.72 &84.02
\\\cline{2-6}
 
& \bf{CCFG-Net}&  \bf{86.43} & \bf{86.34} & \bf{88.37} & \bf{88.04}

\\\hline
\end{tabular}}
\end{center}
\end{table}



\section{Ablation Study}
\label{abs}

In this section, the importance of each component of our approach will be studied. The ablation study presented in Table \ref{tab:abls} investigates the importance of different loss functions in the proposed CCFG-Net approach using Resnet-50 and ViT Base backbones on the CRC-FGVC dataset. The four loss functions considered are ${\mathcal {L}}_{Focal}$, ${\mathcal {L}}_{LMCL}$, ${\mathcal {L}}_{e}$, and ${\mathcal {L}}_{a}$, which correspond to Focal, LMCL, $L_2$ Pairwise contrastive, and Angular Pairwise contrastive losses, respectively.


\begin{table}[ht]
 \caption{Ablation study of the proposed losses for our CCFG-Net approach using ResNet-50 and ViT Base backbones on CRC-FGVC dataset.  \textbf{${\mathcal {L}}_{F}$},  \textbf{${\mathcal {L}}_{L}$}, \textbf{${\mathcal {L}}_{e}$} and \textbf{${\mathcal {L}}_{a}$} correspond to ${\mathcal {L}}_{Focal}$,  ${\mathcal {L}}_{LMCL}$,  ${\mathcal {L}}_{e}$, and  ${\mathcal {L}}_{a}$  in Eq. \ref{totalloss}, respectively. There losses represent Focal , LMCL, $L_2$ Pairwise contrastive and Angular Pairwise contrastive losses, respectively.}
 \begin{center}
\label{tab:abls}
\centering
\scalebox{0.8}{\begin{tabular}{|l|l|cccc|c|c|c|c|}

\hline
   {\multirow{2}{*}    \textbf{Backbone}} &\multirow{2}{*}\textbf{Ex}   &\multicolumn{4}{|c|}{\textbf{Ablation}}  & \multicolumn{2}{|c|}{\textbf{Validation}}& \multicolumn{2}{|c|}{\textbf{Test}} \\
\cline{3-10}


& &\textbf{${\mathcal {L}}_{F}$}&\textbf{${\mathcal {L}}_{e}$} &\textbf{${\mathcal {L}}_{L}$} &\textbf{${\mathcal {L}}_{a}$}      & \textbf{Acc} &\textbf{F1-S} &   \textbf{Acc}& \textbf{F1-S}   \\
\hline

& 1 & \cmark&\xmark & \xmark &  \xmark& 49.17  & 50.51 & 48.20 &49.77 \\ \cline{2-10}

&2 & \cmark&\cmark & \xmark &  \xmark&78.21  & 78.10 &77.27   & 77.67\\\cline{2-10} 

& 3 & \xmark&\xmark & \cmark &  \cmark& 77.93  & 77.51 & 77.11  & 77.16 \\\cline{2-10}

Resnet-50& 4 & \cmark&\xmark & \cmark &  \cmark& 79.05 &  78.43&   78.41& 78.21 \\\cline{2-10}

& 5 & \xmark&\cmark & \cmark &  \cmark&  78.81 & 78.35  & 78.17&78.01 \\\cline{2-10}

& 6  & \cmark&\cmark & \xmark &  \cmark& 78.29& 78.25& 77.15& 77.54\\\cline{2-10}

& 7 & \cmark&\cmark & \cmark &  \xmark&79.94  &79.45  & 78.37  & 78.48\\\cline{2-10}

& 8 & \cmark&\cmark & \cmark &  \cmark&\bf{79.89}  & \bf{79.70} &  \bf{79.20}  & \bf{79.30} \\\hline \hline


& 1 & \cmark&\xmark & \xmark &  \xmark& 58.18   &56.97  &57.87   & 57.30 \\ \cline{2-10}

&2 & \cmark&\cmark & \xmark &  \xmark& 80.14 & 79.91 & 78.82  & 78.77\\\cline{2-10} 

& 3 & \xmark&\xmark & \cmark &  \cmark& 82.90  &  82.60& 82.90  & 82.81 \\\cline{2-10}

ViT& 4 & \cmark&\xmark & \cmark &  \cmark& 83.27 & 82.86 &  82.54 & 82.47 \\\cline{2-10}

(Base)& 5 & \xmark&\cmark & \cmark &  \cmark& 83.72  & 83.30 & 83.10  &82.98 \\\cline{2-10}

& 6  & \cmark&\cmark & \xmark &  \cmark&  79.26 & 78.91 & 78.93  & 79.07\\\cline{2-10}

& 7 & \cmark&\cmark & \cmark &  \xmark&84.24  & 83.75 &  83.32 & 83.14\\\cline{2-10}

& 8 & \cmark&\cmark & \cmark &  \cmark&\bf{84.42} &  \bf{84.09} &  \bf{83.51}  & \bf{83.38}  \\\hline

\end{tabular}}
\end{center}
\end{table}

In Table \ref{tab:abls}, the first experiment serves as the baseline by training the backbone with the Focal loss. The second and third rows examine the results when only the losses of the first head (\({ \mathcal{L}}_{F}\) and \({ \mathcal{L}}_{e}\)) and the losses of the second head (\({ \mathcal{L}}_{LMCL}\) and \({ \mathcal{L}}_{a}\)) are considered, respectively. The results indicate significant improvements achieved by each head loss compared to the baseline. It is also observed that the ResNet-50 backbone performs better with distance-based losses, while the ViT Base backbone shows better performance with angular losses. This finding suggests that merging \(L_2\) Pairwise contrastive losses and Angular Pairwise contrastive losses in each head plays a crucial role in enhancing the model's discriminative power.

Furthermore, analyzing the significance of each loss individually (rows 4 to 7) reveals their respective contributions to performance enhancement compared to the previous ablation studies (rows 2 and 3). However, an exception is observed when removing the ${\mathcal {L}}_{L}$ loss and utilizing only ${\mathcal {L}}_{a}$ in the second head (row 6), which results in a drop in performance compared to using only the losses of the first head. This suggests that ${\mathcal {L}}_{L}$ is essential for stability in the second head, and the Angular Pairwise contrastive loss adds more value by capturing complex relationships between feature vectors and enhancing the model's discriminative capability.

In the final experiment (row 8), incorporating all of the four loss functions (${ \mathcal{L}}_{F}$, ${ \mathcal{L}}_{L}$, ${ \mathcal{L}}_{e}$, and ${ \mathcal{L}}_{a}$) leads to the highest accuracy and F1-Score on both the validation and test sets for both ResNet-50 and ViT Base backbones. This demonstrates that the combination of all loss functions leads to the most effective and discriminative representation learning. Overall, the ablation study highlights the importance of each loss function in improving the model's performance, with the $L_2$ Pairwise contrastive loss and the Angular Pairwise contrastive loss playing significant roles in enhancing discriminative capabilities.

Table \ref{tab:init} investigates the significance of the first training stage and weight initialization using ResNet-50 and ViT-Base backbones on the CRC-FGVC dataset. The baseline experiment establishes the reference point by training the backbone with the CE loss. The subsequent experiments explore different approaches for weight initialization in our CCFG-Net. In the second experiment, the weights are initialized with pretrained ImageNet weights, while in the third experiment, the weights obtained from the training in experiment 1 serve as the initialization. Rows 4 and 5 showcase the performance of SCL \cite{khosla2020supervised} and our CCFG-Net, respectively, when the weights are initialized with SCL weights.  It is important to note that experiment 3 in Table \ref{tab:init} represents the training of our CCFG-Net without the first training stage. On the other hand, the third and fifth experiments correspond to the training of our CCFG-Net with the backbone trained using CE and SCL loss functions in the first training stage, respectively.

Comparing the results of the first and second rows in Table \ref{tab:init}, our architecture demonstrates higher performance in validation and testing data for both ResNet-50 and ViT Base backbones when the backbone is initialized with ImageNet pretrained weights. Notably, ViT Base shows a larger improvement margin compared to its baseline, with a testing data F1-score improvement of 14.82\% compared to 8.74\% for ResNet-50. Furthermore, the third experiment highlights the significance of the first training stage, resulting in a substantial testing data F1-score improvement of 17.39\% for ResNet-50 and 20.56\% for ViT Base compared to the baseline. The last two experiments further support the effectiveness of our proposed CCFG-Net, demonstrating improved performance when the backbone is trained with SCL in the first stage. In conclusion, these ablation studies emphasize the importance of proper initialization and the inclusion of the first training stage in enhancing the performance of CCFG-Net.

\begin{table}[ht]
 \caption{Ablation Study of the Proposed Approach with Different First Training Stage and Initialization for ResNet-50 and ViT Base Backbones. }
 \begin{center}
\label{tab:init}
\centering
\scalebox{0.7}{\begin{tabular}{|l|l|l|l|c|c|c|c|}

\hline
  {\multirow{2}{*}    \textbf{Backbone} }      &  \multirow{2}{*}\textbf{Ex}&{\multirow{2}{*}    \textbf{Method}}& {\multirow{2}{*}    \textbf{Initialization} }  & \multicolumn{2}{|c|}{\textbf{Validation}}& \multicolumn{2}{|c|}{\textbf{Test}}\\
\cline{4-8}


 & & & & \textbf{Acc} &\textbf{F1-S} &   \textbf{Acc}& \textbf{F1-S} 
 \\
\hline

& 1& CE loss & ImageNet &50.28 & 51.50 & 49.09  & 50.94 \\\cline{2-8}
& 2& CCFG-Net &ImageNet& 59.70  & 59.49 & 59.03  & 59.68
  \\\cline{2-8}

Resnet-50& 3&CCFG-Net & CE Trained Model& 68.56  & 68.04 & 68.67  & 68.33
\\\cline{2-8}

 &  4&SCL loss \cite{khosla2020supervised} & ImageNet&  74.82 & 74.56 &  74.77 & 74.96
  \\\cline{2-8}

& 5&\bf{CCFG-Net} &SCL Trained Model & \bf{79.89}  & \bf{79.70} &  \bf{79.20}  & \bf{79.30}  

\\\hline\hline

& 1&CE loss& ImageNet&58.18   &56.97  &57.87   & 57.30
  \\\cline{2-8} 
& 2&CCFG-Net &ImageNet& 74.13  &73.49  & 72.44  & 72.12
  \\\cline{2-8}

ViT (Base) & 3&CCFG-Net &CE Trained Model&  78.73 & 78.06 & 77.96  & 77.86
  \\\cline{2-8}

&4  &SCL loss \cite{khosla2020supervised}& ImageNet& 76.53  & 75.88 & 74.72  & 74.76
   \\\cline{2-8}

& 5& \bf{CCFG-Net }& SCL Trained Model&\bf{84.42} &  \bf{84.09} &  \bf{83.51}  & \bf{83.38} 

\\\hline

\end{tabular}  }
\end{center}
\end{table}

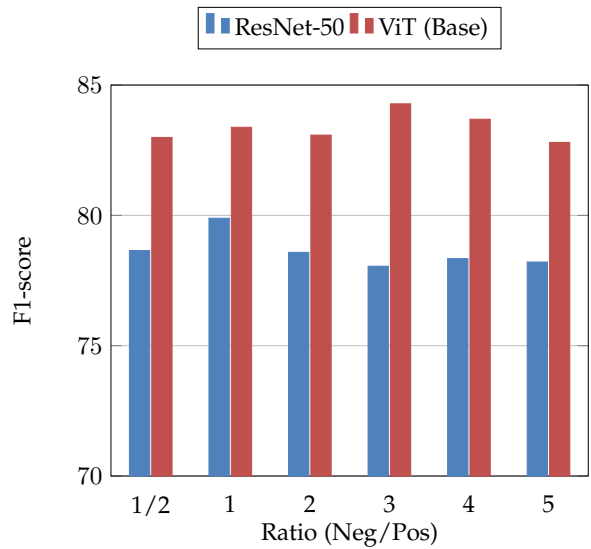
\begin{figure}[ht]
\centering
\scalebox{0.96}{
\begin{tikzpicture}
    \begin{axis}[
        width  = 0.45*\textwidth,
        height = 7cm,
        major x tick style = transparent,
        ybar=2*\pgflinewidth,
        bar width=8pt,
        ymajorgrids = true,
        xlabel = {Ratio (Neg/Pos)},
        ylabel = {F1-score},
        symbolic x coords={1/2,1,2, 3, 4, 5},
        xtick = data,
        scaled y ticks = false,
        enlarge x limits=0.10,
        ymin=70,
        ymax=85,
        legend style={
            at={(0.5,1.2)},
            anchor=north,
            legend columns=2,
            /tikz/every even column/.append style={column sep=0cm}
        },
    ]
        \addplot[style={bblue,fill=bblue,mark=none}]
             coordinates {(1/2,78.65) (1,79.89) (2,78.58) (3,78.05) (4, 78.34) (5,78.21)};

        \addplot[style={rred,fill=rred,mark=none}]
             coordinates {(1/2,82.99) (1,83.38) (2,83.08) (3,84.28) (4,83.69) (5,82.80)};

        \legend{ResNet-50, ViT (Base)}
    \end{axis}
\end{tikzpicture}}
\caption{Study of the Negative Pairs Ratio to Positive Pairs for ResNet-50 and ViT Base Backbones. }
\label{fig:np}
\end{figure}

Our approach utilizes two training stages, as described in Section \ref{prapp}. In the second training stage, all possible positive pairs are selected to construct the batches, while a random selection of $p_n$ negative classes is made for each image in the training data to form negative pairs for each epoch. Figure \ref{fig:np} illustrates the results of the proposed CCFG-Net for different ratios of $Neg/Pos$ using ResNet-50 and ViT Base backbones on the CRC-FGVC dataset. The figure reveals that a ratio of 1 achieves the best performance for ResNet-50 backbone. This indicates that giving equal importance to negative and positive pairs in CCFG-Net (ResNet-50) strikes a favorable balance between performance and training time, as including more negative pairs would increase training time. However, for ViT Base backbone, the optimal ratio is 3. This suggests that transformers benefit from more training data, but there is a point where a balance between negative and positive pairs becomes necessary. In our approach, we adopt an equalization scenario between positive and negative pairs for all backbones, achieving a trade-off between performance and training time.

\begin{figure}[ht]
\centering
\includegraphics[width=9cm, trim={3cm 1cm 3cm 2cm},clip]{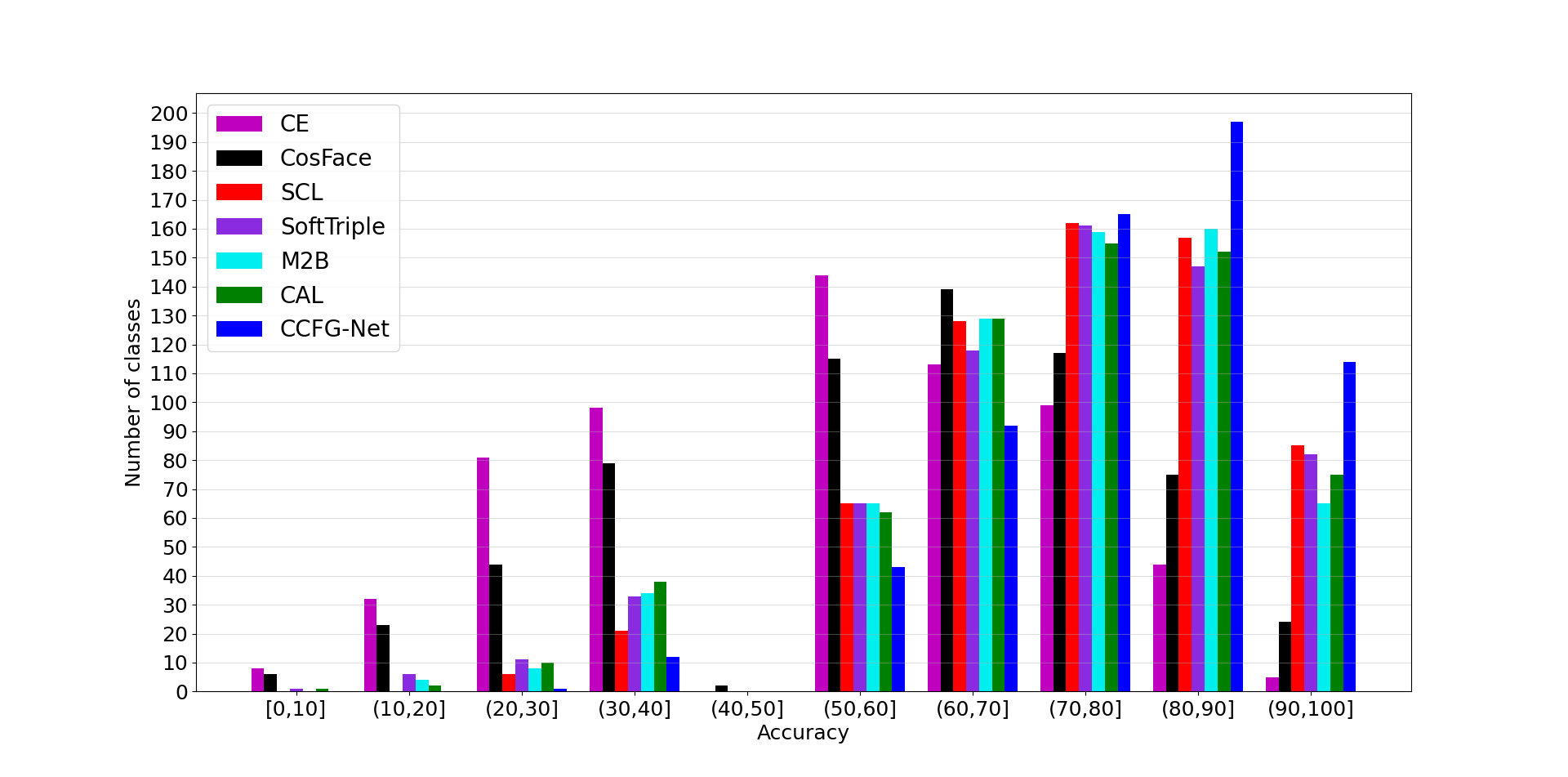}
\caption{Accuracy Histogram  Comparison of ResNet-50  Backbones on RCC-FGVC dataset. } 
\label{fig:histres}
\end{figure}

\section{Discussion} 
\label{discu}
In this section, we analyze the results obtained from our experiments. Fig. \ref{fig:histres} visualizes the distribution of class accuracies for our proposed CCFG-Net and six other approaches, all using the ResNet-50 backbone on the CRC-FGVC testing data. The CE (baseline) and CosFace approaches show a normal distribution centered around 50-60\% accuracy. In contrast, the other approaches, including ours, exhibit a right-skewed distribution. Notably, CCFG-Net has the highest mean accuracy and the smallest median, followed by SCL and CAL. Specifically, for CCFG-Net, 20\% of classes achieve 90-100\% accuracy, one-third fall within 80-90\%, and one-fourth between 70-80\%. These results underscore the effectiveness of CCFG-Net in discriminating Chinese characters.



\begin{CJK*}{UTF8}{gbsn}

Tables \ref{tab:exmp1} and \ref{tab:exmp2} present challenging samples from the RCC-FGVC testing data and the corresponding predictions from our proposed approach and six comparison methods, utilizing ResNet-50 and ViT Base backbones, respectively. These examples highlight the diverse challenges in real-world scenarios. For ResNet-50 samples, challenges include resembling characters, variations in storefront signage design (e.g., examples 1 and 3), inclusion of extraneous parts during detection and cropping (e.g., examples 2 and 4), and low-quality images due to blurriness, distance, or environmental effects (e.g., examples 5 to 7). Background and writing color can also make distinguishing character details difficult. In Table \ref{tab:exmp1}, our approach correctly classifies the first nine examples, overcoming these challenges, while comparison methods often misclassify these samples as resembling classes. For the last two examples, all methods fail, although our approach only misses tiny details, suggesting potential improvements with augmented training data.

Similarly, in Table \ref{tab:exmp2} for ViT Base samples, we observe similar challenges affecting the comparison methods' performance. These challenges include resembling characters, real-world conditions, background effects, low image quality, and variations in advertising style. Despite these difficulties, our approach effectively classifies these examples. This success is due to our proposed compound loss function, which enables the comparison of positive and negative pairs, facilitating the identification of tiny distinguishable details between characters while learning their general structure within the deep representation. Consequently, our approach eliminates the need for handcrafted methods, which can exhibit inconsistent performance across different tasks, datasets, and conditions.


\begin{table}[ht]
 \caption{Misclassified Samples ResNet-50 Backbone. }
 \begin{center}
\label{tab:exmp1}
\centering
\scalebox{0.5}{\begin{tabular}{c|c|ccccccc|c}

   \textbf{Num}    & \backslashbox{Example}{Approach}    &   \textbf{CE}& \textbf{CosFace}& \textbf{SCL} & \textbf{SoftTripple} & \textbf{M2B} & \textbf{CAL} &\textbf{CCFG-Net} & \textbf{GT}\\

\hline

1&\begin{minipage}{.2\textwidth}
\includegraphics[width=\linewidth, height=20mm]{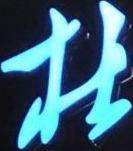}
\end{minipage}&  {\Huge 姓}
 &  {\Huge 进}
  &  {\Huge 杜}
 &  {\Huge 社}
 &  {\Huge 社}
 &  {\Huge 桂}
 &  {\Huge 杜}
  &  {\Huge 杜}

\\

2&\begin{minipage}{.2\textwidth}
\includegraphics[width=\linewidth, height=20mm]{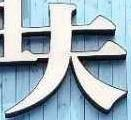}
\end{minipage}& {\Huge 大}
 & {\Huge大}
  & {\Huge 大}
 & {\Huge 夫 }& {\Huge 大}  &{\Huge 夹}
 & {\Huge 夫}
  & {\Huge 夫}
 \\

3&\begin{minipage}{.2\textwidth}
\includegraphics[width=\linewidth, height=20mm]{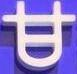}
\end{minipage}&{\Huge 时}
 & {\Huge 时 }
  & {\Huge 胡}
 & {\Huge 百}
 & {\Huge 甘}
  & {\Huge 时}
 & {\Huge 甘}
  & {\Huge 甘}
 \\

4&\begin{minipage}{.2\textwidth}
\includegraphics[width=\linewidth, height=20mm]{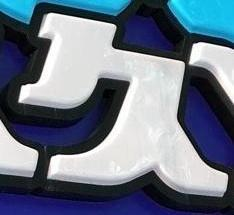}
\end{minipage}&{\Huge 久}
 &{\Huge 多}
  & {\Huge 人}
 & {\Huge 久}
&{\Huge 又}
  & {\Huge 又}
 & {\Huge 久}
  & {\Huge 久}
 \\

5&\begin{minipage}{.2\textwidth}
\includegraphics[width=\linewidth, height=20mm]{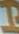}
\end{minipage} &{\Huge 工}
 & {\Huge 工}
  & {\Huge 工}
 & {\Huge 工}
 & {\Huge 上}
  & {\Huge 卫}
 & {\Huge 卫}
  & {\Huge 卫}
 
 \\

6&\begin{minipage}{.2\textwidth}
\includegraphics[width=\linewidth, height=20mm]{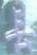}
\end{minipage}& {\Huge县}
 &  {\Huge县}  & {\Huge 县}
 & {\Huge具}

 &  {\Huge具}
  & {\Huge具}
 & {\Huge 县}
  &  {\Huge县}

\\

7&\begin{minipage}{.2\textwidth}
\includegraphics[width=\linewidth, height=20mm]{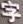}
\end{minipage}& {\Huge车}
 &  {\Huge字} &  {\Huge宇} & {\Huge 字} &  {\Huge字}  &  {\Huge宇} & {\Huge字}  & {\Huge 字}

\\

8&\begin{minipage}{.2\textwidth}
\includegraphics[width=\linewidth, height=20mm]{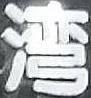}
\end{minipage}& {\Huge披} &  {\Huge地}  & {\Huge 沙}
 & {\Huge 汕} & {\Huge 汕} &  {\Huge汕} & {\Huge 湾} & {\Huge 湾}

\\

9&\begin{minipage}{.2\textwidth}
\includegraphics[width=\linewidth, height=20mm]{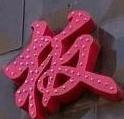}
\end{minipage}& {\Huge饭} &  {\Huge饭}  &  {\Huge枝}
 & {\Huge 校} &  {\Huge饭}  &  {\Huge饭} & {\Huge板}   & {\Huge 板}

\\\hline
10&\begin{minipage}{.2\textwidth}
\includegraphics[width=\linewidth, height=20mm]{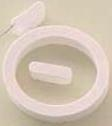}
\end{minipage}  & {\Huge古} &  {\Huge古}  &  {\Huge自} &  {\Huge宫} &  {\Huge简}  &  {\Huge节} &  {\Huge自}  &  {\Huge白}
 
\\
11&\begin{minipage}{.2\textwidth}
\includegraphics[width=\linewidth, height=20mm]{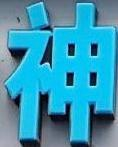}
\end{minipage}& {\Huge 为}
 &  {\Huge 科 } &  {\Huge 唯} &  {\Huge 名} & {\Huge 湾}  &  {\Huge 卤} & {\Huge 浩}  & {\Huge 神 }

\\

\end{tabular}}
\end{center}
\end{table}



\begin{table}[ht]
 \caption{Misclassified Samples ViT Base Backbone. }
 \begin{center}
\label{tab:exmp2}
\centering
\scalebox{0.5}{
\begin{tabular}{c|c|ccccccc|c}

  \textbf{Num}    & \backslashbox{Example}{Approach}    &   \textbf{CE}& \textbf{CosFace}& \textbf{SCL} & \textbf{SoftTripple} & \textbf{M2B} & \textbf{CAL} &\textbf{CCFG-Net} & \textbf{GT}\\

\hline


1&\begin{minipage}{.2\textwidth}
\includegraphics[width=\linewidth, height=20mm]{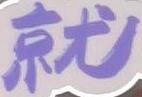}
\end{minipage}&{\Huge 晓}
 &{\Huge 就}
  & {\Huge 就}
 & {\Huge 锅}
&{\Huge 树}
  & {\Huge 就}
 & {\Huge 就}
  & {\Huge 就}
 \\

2&\begin{minipage}{.2\textwidth}
\includegraphics[width=\linewidth, height=20mm]{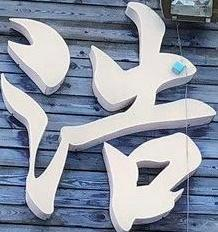}
\end{minipage}& {\Huge 活}
 & {\Huge  诺}
  & {\Huge 浩}
 & {\Huge 话}& {\Huge 湖}  &{\Huge 湖}
 & {\Huge 浩}
  & {\Huge 浩}
 \\

3&\begin{minipage}{.2\textwidth}
\includegraphics[width=\linewidth, height=20mm]{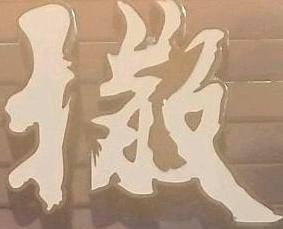}
\end{minipage}&{\Huge 烧}
 & {\Huge 徽}
  & {\Huge 微}
 & {\Huge 馍}
 & {\Huge 铁}
  & {\Huge 缘}
 & {\Huge 徽}
  & {\Huge 徽}
 \\

4&\begin{minipage}{.2\textwidth}
\includegraphics[width=\linewidth, height=20mm]{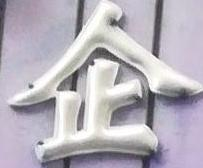}
\end{minipage}&{\Huge 企} &{\Huge 企}  & {\Huge 金}
 & {\Huge 金}
&{\Huge 金}  & {\Huge 金} & {\Huge 企}  & {\Huge 企}
 \\

5&\begin{minipage}{.2\textwidth}
\includegraphics[width=\linewidth, height=20mm]{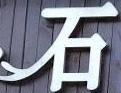}
\end{minipage} &{\Huge 万}
 & {\Huge 石 }
  & {\Huge 石}
 & {\Huge 万}
 & {\Huge 都}
  & {\Huge 万}
 & {\Huge 石}
  & {\Huge 石}
 
 \\

6&\begin{minipage}{.2\textwidth}
\includegraphics[width=\linewidth, height=20mm]{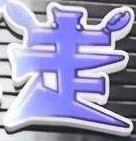}
\end{minipage}& {\Huge 主} &  {\Huge 注}  & {\Huge 注} & {\Huge 走} &  {\Huge 主} & {\Huge 麦} & {\Huge  走}  &  {\Huge 走}

\\

7&\begin{minipage}{.2\textwidth}
\includegraphics[width=\linewidth, height=20mm]{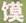}
\end{minipage}& {\Huge 域}
 &  {\Huge 设} &  {\Huge 道} & {\Huge 间 } &  {\Huge 健}  &  {\Huge 限} & {\Huge 馍}  & {\Huge 馍}
 
\\

8&\begin{minipage}{.2\textwidth}
\includegraphics[width=\linewidth, height=20mm]{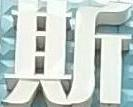}
\end{minipage}&  {\Huge 新}
 &  {\Huge 折} &  {\Huge 燕} & {\Huge  事} &  {\Huge 新} 
 &  {\Huge  燕} & {\Huge  斯}  & {\Huge  斯}

\\

8&\begin{minipage}{.2\textwidth}
\includegraphics[width=\linewidth, height=20mm]{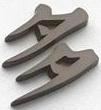}
\end{minipage}&  {\Huge 室}
 &  {\Huge 经} &  {\Huge 重 } & {\Huge 重} &  {\Huge 重}  
 &  {\Huge  重} & {\Huge  多}  & {\Huge  多}
 
\\
9&\begin{minipage}{.2\textwidth}
\includegraphics[width=\linewidth, height=20mm]{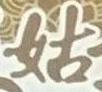}
\end{minipage}  & {\Huge 居}
 &  {\Huge 范} &  {\Huge 转} & {\Huge  转} &  {\Huge 晓}  &  
 {\Huge  胡} & {\Huge  姑}  & {\Huge  姑}
 
\\\hline
10&\begin{minipage}{.2\textwidth}
\includegraphics[width=\linewidth, height=20mm]{3068.png}
\end{minipage}&  {\Huge 潮}
 &  {\Huge 潮} &  {\Huge 潮} & {\Huge 潮} &  {\Huge 潮}  
 &  {\Huge  潮} & {\Huge  潮}  & {\Huge  神}

\\

\end{tabular}}
\end{center}
\end{table}
\end{CJK*}


\section{Conclusion}
\label{conc}
In this paper, we tackle the challenge of recognizing resembling glyphs in the wild by constructing the RCC-FGVC dataset, based on a proposed resembling dictionary, and additionally provided an English letters dataset, EL-FGVC. We introduced a novel two-stage Siamese contrastive learning approach, CCFG-Net, designed to enhance character recognition. Our approach demonstrated significant effectiveness across five backbones, including both CNN and Transformer architectures.
Comprehensive evaluations and comparisons with state-of-the-art fine-grained classification methods highlighted the superiority of our proposed method for recognizing resembling Chinese characters and English letters in natural scenes. Specifically, our method effectively leverages supervised contrastive learning and integrates classification and contrastive learning in both Euclidean and angular spaces, leading to improved discriminative power.
This work not only addresses the recognition problem in the wild from few-shot samples but also provides a challenging benchmark for evaluating fine-grained approaches. Despite the progress achieved, further investigation is needed to construct a more comprehensive resembling dictionary encompassing a wider range of characters and better resembling groups. Additionally, enhancing the current approach to consider resembling classes during training, rather than randomly selecting negative classes to reconstruct pairs, could offer better performance and improved handling of resembling challenges.


%



\ifCLASSOPTIONcompsoc
 \section*{Acknowledgments}
\else
   regular IEEE prefers the singular form
 \section*{Acknowledgment}
\fi

The corresponding author was partially supported by Henan Provincial Science and Technology Project (No.232102211021) and MOE Liberal Arts and Social Sciences Foundation (No.23YJAZH210). 

\ifCLASSOPTIONcaptionsoff
  \newpage
\fi

\end{document}